\documentclass[acmsmall]{acmart}

\setcopyright{acmcopyright}
\acmJournal{PACMHCI}
\acmYear{2022} \acmVolume{6} \acmNumber{ETRA} \acmArticle{139}
\acmMonth{5} \acmPrice{15.00}\acmDOI{10.1145/3530880}

\AtBeginDocument{%
  \providecommand\BibTeX{{%
    \normalfont B\kern-0.5em{\scshape i\kern-0.25em b}\kern-0.8em\TeX}}}

\usepackage{cancel}
\usepackage{layout}
\usepackage{caption}
\usepackage{diagbox}
\usepackage{natbib}
\usepackage{graphicx}
\usepackage{multirow}
\usepackage{nicefrac}
\usepackage{subfiles}
\usepackage{hyperref}
\usepackage{arydshln}
\usepackage{placeins}
\usepackage{mathtools}
\usepackage{adjustbox}
\usepackage{changepage}
\usepackage{subcaption}
\usepackage{soul,xcolor}

\newcommand{\etal}{\textit{et al}. }
\newcommand{\ie}{\textit{i}.\textit{e}., }

\newcommand{\rev}[1]{\color{black} #1 \color{black}}

\begin{document}

\title{EllSeg-Gen, towards Domain Generalization for head-mounted eyetracking}

\author{Rakshit S. Kothari}
\email{rsk3900@rit.edu}
\orcid{0000000243183068}
\authornotemark[1]
\email{rsk3900@rit.edu}
\affiliation{%
    \institution{Rochester Institute of Technology}
    \streetaddress{1 Lomb Memorial Drive}
    \city{Rochester}
    \state{NY}
    \country{USA}
\postcode{14623}
}

\author{Reynold J. Bailey}
\affiliation{%
    \institution{Rochester Institute of Technology}
    \streetaddress{1 Lomb Memorial Drive}
    \city{Rochester}
    \state{NY}
    \country{USA}
\postcode{14623}
}

\author{Christopher Kanan}
\affiliation{%
    \institution{Rochester Institute of Technology}
    \streetaddress{1 Lomb Memorial Drive}
    \city{Rochester}
    \state{NY}
    \country{USA}
\postcode{14623}
}

\author{Jeff B. Pelz}
\affiliation{%
    \institution{Rochester Institute of Technology}
    \streetaddress{1 Lomb Memorial Drive}
    \city{Rochester}
    \state{NY}
    \country{USA}
\postcode{14623}
}

\author{Gabriel J. Diaz}
\affiliation{%
    \institution{Rochester Institute of Technology}
    \streetaddress{1 Lomb Memorial Drive}
    \city{Rochester}
    \state{NY}
    \country{USA}
\postcode{14623}
}

\renewcommand{\shortauthors}{Rakshit S. Kothari et al.} 

\begin{abstract}
The study of human gaze behavior in natural contexts requires algorithms for gaze estimation that are robust to a wide range of imaging conditions. However, algorithms often fail to identify features such as the iris and pupil centroid in the presence of reflective artifacts and occlusions. Previous work has shown that convolutional networks excel at extracting gaze features despite the presence of such artifacts. However, these networks often perform poorly on data unseen during training. This work follows the intuition that jointly training a convolutional network with multiple datasets learns a generalized representation of eye parts. We compare the performance of a single model trained with multiple datasets against a pool of models trained on individual datasets. Results indicate that models tested on datasets in which eye images exhibit higher appearance variability benefit from multiset training. In contrast, dataset-specific models generalize better onto eye images with lower appearance variability. 
\end{abstract}



\maketitle

\section{Introduction}
Eye tracking solutions frequently employ computer vision or machine learning (ML) algorithms to extract features of interest from images captured using eye cameras. These features facilitate the estimation of a subject’s gaze position. While numerous efforts have explored both approaches, recent works~\cite{Chaudhary2019b,kothari2021ellseg,Yiu2019DeepVOG:Learning,Vera-Olmos2018DeepEye:Environments,Fuhl2019500000Segmentation,Garbin2019OpenEDS:Dataset,kansal2019eyenet} report that ML systems demonstrate state of the art performance in identifying gaze relevant features for head-mounted eyetracking. Contrary to computer vision approaches where features are identified using handcrafted algorithms and heuristics, superior performance by ML is partly achieved by making minor adjustments in a ML system’s internal parameters with the objective to maximize the probability of predicting known outputs for given training inputs~\cite{vapnik1999overview}. A ML system with millions of parameters could theoretically demonstrate perfect performance over the distribution of data it was trained on. These systems however often fail to generalize to out-of-distribution samples that are dissimilar to ones seen during training but plausible under the overarching goals of the problem. For example, ML systems trained to segment eye images acquired with a small subset of the human population, or optimized for particular imaging hardware, may fail to generalize onto the average use case.

\begin{figure}
    \centering
    \includegraphics[width=0.55\linewidth]{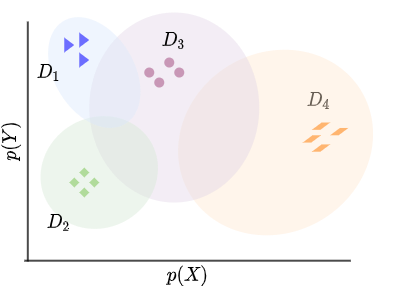}
    \caption{An illustration to visualize two different training strategies one may adopt to maximize generalization. If $D_3$ represents a test distribution, then it is intuitive to train a model using a combination of $D_1$, $D_2$ and $D_4$. However, if $D_4$ represents a test distribution then a model specific to $D_3$ would demonstrate optimal generalization.}
    \label{fig:motiv_multiset}
\end{figure}

In the context of eyetracking, high performance across subjects, environmental reflection, camera quality, camera placement and eye occluding artifacts can be cast as a Domain Generalization problem~\cite{torralba2011unbiased,li2017deeper,Muandet2013,ben2010theory}. Although the intuitive thought is that a broader training set will always produce the most generalizable model, it may come at the cost of performance~\cite{yang2020closer,zhang2019theoretically,raghunathan2020understanding} as a significantly broader distribution must be captured by limited network complexity~\cite{madry2017towards}. A relevant and alternative approach to solving the generalization problem would be to assemble a suite of domain-specific models. These specialized models theoretically have attained the ceiling performance on a domain while remaining agnostic to others. At run-time, one could select a model trained on a domain that maximizes statistical similarity to the intended testing data. The notable drawback to this approach is that it assumes the existence of a hypothetical method which finds the best matching model without any test-time labels or annotations (for example, one may consider model uncertainty~\cite{Kendall2017} as a measure of performance). Figure~\ref{fig:motiv_multiset} is a graphical example to illustrate these approaches one may adopt for optimal generalization.

Collecting and annotating data is a difficult and time consuming task. In context to head-mounted eyetracking, we often collect and annotate a small subset of data drawn from a few users with the hope that our model would generalize to a broader population. Models trained on a biased sampling of the intended distribution may not generalize well beyond a few subjects or gaze positions~\cite{nair2020rit}. For example, a model trained on eye images with a narrow gaze distribution or specific eye camera orientation may not generalize to users with non-overlapping or broader gaze distribution and camera orientation. We hypothesize that jointly training with multiple datasets may expand the available training distribution and in turn improve generalization across data collected from different subjects under the same environmental conditions.

In this work, we explore the relationship between model generalizability and performance within the context of eye tracking. The specific contributions of this work are as follows:
\begin{itemize}
    % If this hypothesis is true
    \item We test the hypothesis that a single model trained on data drawn from multiple heterogeneous domains can generalize better than a specialized, dataset-specific model when evaluated on domains \textbf{unseen during training}. If we accept this hypothesis, then it suggests there is a benefit in expanding the breadth of the training distribution by accumulating more data. If we reject this hypothesis, then the better approach to achieve generalizability is by exploring test-time techniques to find the optimal model from a pool of dataset-specific models.
    
    \item A model may not generalize well across subjects due to a distribution shift caused from a limited or biased sampling of the data distribution. We hypothesize that training with multiple heterogeneous distributions could improve ceiling performance by mitigating distribution shift. If we accept this hypothesis, then it suggests that one may adopt a multiset training approach to identify limitations in a dataset. If we reject this hypothesis, then we are faced with two possibilities. The first possibility is that multiset training exhibits near ceiling performance, suggesting that the dataset has been optimally sampled and the inclusion of a broad training distribution does not aid generalization to new subjects. The second possibility is that multiset training exhibits deteriorated performance, suggesting that the network has limited complexity.
    
    \item In addition to the central hypothesis related to the principles of model generalization, we present multiple contributions and insights of practical use to the eye tracking community. These include analyzing the effects of data augmentation and model complexity.
    
\end{itemize}

\section{Related work}

Issues related to domain generalization arise when a ML model trained on a particular domain generally does not perform well on out-of-domain samples~\cite{ponce2006dataset,torralba2011unbiased,Khosla2012,vapnik1999overview,kouw2018introduction}. This well known phenomenon is known as distribution shift and generally occurs when the distribution of data points used to train the network does not match the distribution of data we are evaluating the model on. There are two major types forms of distributions shifts, prior and covariate shifts.

Prior shift occurs when the distribution of output labels between the train and test domains vary~\cite{kouw2018introduction,kouw2019review}. Consider the case of eye image segmentation, a critical process in head-mounted eyetracking. In this context, a class could represent an annotated eye part at the pixel level (e.g. pupil, iris, sclera, other). A change in class proportions could occur when the eye camera is displaced to different distances or orientations not considered during training. To measure and mitigate prior shift, we require access to a small portion of labels from the test domain. Access to test labels allows us to selectively draw samples during training which better represent the test domain. Domain generalization however assumes no access to test labels. Data augmentation schemes such as artificially translating, rotating and scaling images from a domain can mitigate a portion of prior shift~\cite{wang2017effectiveness,shorten2019survey,van2001art,tanner1987calculation,Park2018LearningSettings}.

A mismatch in eye camera pose or gaze distributions is further convoluted by variation in eye image appearance. Appearances can widely vary depending on the environment, occlusions, subject physiology, pupil dilation, optical aberrations or reflections. Therefore, statistics related to the spatial distribution of group membership at the pixel level is unlikely to match across training/testing domains that reflect a difference in eye camera quality, camera intrinsic parameters, or across different populations and environments. Such a shift in distribution due to discordant appearances results in covariate shift. For example, a model trained to segment eye images captured indoors may not generalize well onto outdoor environments despite enforcing other variables such a subject physiology, gaze position and camera pose to be fixed. There are numerous approaches which attempt to minimize covariate shift but they require access to target imagery. Domain generalization assumes no access to target imagery.

\subsection{Mitigating distribution shift}
%Empirical Risk Minimization~\cite{vapnik1999overview} is one of the earliest algorithms which serves as baseline for generalization. Simply stated, simultaneous optimization across multiple domains results in a model with the least empirical risk on a broader distribution. While simple in design, multiset training serves as a baseline to measure other techniques for domain generalization.

Previous work in domain generalization follow the intuition that a model may generalize better if internal network activations are invariant across domain-specific factors~\cite{torralba2011unbiased,Khosla2012,ghifary2015domain}. In the context of eyetracking, we want to encourage a network to learn a generalized representation of an eye image and its subsequent mapping to semantic categories which are invariant to camera quality, occluding artifacts, gaze position and eye camera location. This is achieved by penalizing a network when the learned latent representation of eye images statistically align themselves to the domain they were sampled from. Numerous attempts have been made which attempt to align representations across domains. Approaches involve leveraging the Mean Discrepancy metric~\cite{borgwardt2006integrating,tzeng2014deep,li2018domain}, curriculum based learning~\cite{huang2020episodic}, meta-learning~\cite{Balaji2018,li2018learning,finn2017model} and adversarial learning~\cite{shen2020domain,tzeng2017adversarial,li2018domain} using min-max optimization~\cite{goodfellow2014generative} or gradient reversal~\cite{ganin2015unsupervised}.

While these techniques have demonstrated better generalization performance in their respective applications, their effectiveness for head-mounted eye image segmentation remains unclear. Extensive experiments by Gulrajani~\etal~\cite{Gulrajani2020} reveal that one of the earliest and simplest approaches, Empirical Risk Minimization (ERM)~\cite{vapnik1999overview} results in similar or better performance as compared to state of the art approaches when implemented and evaluated correctly. Simply stated, ERM involves optimizing a model to minimize its risk of misclassification simultaneously across multiple domains. In this work, we refer to ERM as multiset training. Due to its simple and intuitive design, multiset training often serves as a baseline to measure other techniques for domain generalization. This work adopts multiset training to explore two hypotheses briefly introduced in the previous section by leveraging insights from Gulrajani~\etal

\section{Methods}

We describe the datasets, encoder-decoder convolutional architecture and various tests which allow us to test our hypothesis that jointly optimizing using multiple datasets results in better generalization than finding an optimal dataset to train a model.

\subsection{Datasets}

Tests of generalizability are complicated by practical limitations in dataset acquisition. Acquiring data which represents the general populace requires large-scale data collection with annotated ground truth. This needs many hours of tedious work, careful labelling, and may be impractical without the investment of significant time, financial resources, and coordinated effort across multiple laboratories. To mitigate this limitation, this work exploits multiple pre-existing and publicly available datasets of near-eye images acquired from a large number of subjects and different eye trackers which are captured under varying environmental conditions.

We acquire eye imagery from nine publicly available, annotated and heterogeneous datasets (see Table~\ref{tab:dataset_avail} for overall statistics, Supplementary Table 1 for dataset splitting procedure and Supplementary Figure 2 for individual distribution plots). Eye images from each dataset are assigned to a \textit{train} or \textit{test} split. Each split contains images acquired from different subjects or recording IDs. Images present in each dataset are scaled to a common resolution of 320 $\times$ 240 pixels. OpenEDS eye images were cropped vertically to maintain a constant aspect ratio across all datasets. Cropping was performed in a manner which ensured that the entire iris ellipse is visible. Note that the sets published in ExCuSe~\cite{fuhl2015excuse}, ElSe~\cite{Fuhl2016ElSe:Environments} and PupilNet~\cite{Fuhl2017PupilNetDetection} are combined into a single dataset as the source of eye imagery are from the same collection~\cite{kasneci2014driving}. We refer to the combination of these datasets as the \textit{Fuhl} datasets.

\begin{table}[]
\begin{adjustwidth}{-.0cm}{0cm}
\centering
\renewcommand{\arraystretch}{1.25}
% \begin{adjustbox}{angle=90}
\begin{tabular}{|c|c|c|c|c|c|c|c|} 
\hline
Cond                         & Res             & Status & Dataset                                                                       & \multicolumn{2}{c|}{Subject ID}                                                                                                                                                                                                                              & \multicolumn{2}{c|}{\begin{tabular}[c]{@{}c@{}}\# of Images\\(Subjects)\end{tabular}}                      \\ 
\hline
                             &                 &        &                                                                               & Train                                                                                                                                & Test                                                                                                                  & Train                                               & Test                                                 \\ 
\hline
\multirow{6}{*}{\rotatebox[origin=c]{90}{Constrained}} & 640$\times$480  & Synth  & S-General                                                                     & 1 - 18                                                                                                                               & 19 - 24                                                                                                               & \begin{tabular}[c]{@{}c@{}}34254\\(18)\end{tabular} & \begin{tabular}[c]{@{}c@{}}11582\\(6)\end{tabular}   \\ 
\cdashline{2-8}
                             & 400$\times$640  & Real   & OpenEDS'19                                                                    & Train set                                                                                                                            & Valid set                                                                                                             & \begin{tabular}[c]{@{}c@{}}8827\\(95)\end{tabular}  & \begin{tabular}[c]{@{}c@{}}2386\\(28)\end{tabular}   \\ 
\cdashline{2-8}
                             & 1280$\times$960 & Synth  & NVGaze                                                                        & \begin{tabular}[c]{@{}c@{}}Male 1-4,\\ Female 1-4\end{tabular}                                                                       & \begin{tabular}[c]{@{}c@{}}Male 5,\\ Female 5\end{tabular}                                                            & \begin{tabular}[c]{@{}c@{}}16000\\(8)\end{tabular}  & \begin{tabular}[c]{@{}c@{}}4000\\(2)\end{tabular}    \\ 
\cdashline{2-8}
                             & 640$\times$480  & Real   & LPW$^*$                                                                       & \begin{tabular}[c]{@{}c@{}}2,4,5,7,10,\\ 11,13,14,17,\\ 19,21,22\end{tabular}                                                        & \begin{tabular}[c]{@{}c@{}}3,6,8,9,12,\\ 15,16,18,20\end{tabular}                                                     & \begin{tabular}[c]{@{}c@{}}24000\\(12)\end{tabular} & \begin{tabular}[c]{@{}c@{}}17730\\(10)\end{tabular}  \\ 
\cdashline{2-8}
                             & 640$\times$480  & Real   & Swirski                                                                       & 1                                                                                                                                    & 2                                                                                                                     & \begin{tabular}[c]{@{}c@{}}298\\(1)\end{tabular}    & \begin{tabular}[c]{@{}c@{}}298\\(1)\end{tabular}     \\ 
\cdashline{2-8}
                             & 384$\times$288  & Real   & BAT                                                                           & 1,2,3                                                                                                                                & 4,5,6                                                                                                                 & \begin{tabular}[c]{@{}c@{}}3662\\(3)\end{tabular}   & \begin{tabular}[c]{@{}c@{}}3541\\(3)\end{tabular}    \\ 
\hline
\multirow{3}{*}{\rotatebox[origin=c]{90}{Outdoors}}    & 640$\times$480  & Synth  & S-Natural                                                                     & 1 - 18                                                                                                                               & 19 - 24                                                                                                               & \begin{tabular}[c]{@{}c@{}}34267\\(18)\end{tabular} & \begin{tabular}[c]{@{}c@{}}11548\\(6)\end{tabular}   \\ 
\cdashline{2-8}
                             & 640$\times$480  & Synth  & UnityEyes                                                                     & -                                                                                                                                    & -                                                                                                                     & \begin{tabular}[c]{@{}c@{}}16000\\(-)\end{tabular}  & \begin{tabular}[c]{@{}c@{}}2000\\(-)\end{tabular}    \\ 
\cdashline{2-8}
                             & 384$\times$288  & Real   & \begin{tabular}[c]{@{}c@{}}(Fuhl)\\ ExCuSe +\\ ElSe +\\ PupilNet\end{tabular} & \begin{tabular}[c]{@{}c@{}}I, III, V, VII,\\ VIII, XI, XII,\\ XIV, XVI, XVIII,\\ XIX, XX, XXI,\\ XXIV,\\ New II, New IV\end{tabular} & \begin{tabular}[c]{@{}c@{}}II, IV, VI, IX,\\ X, XIII, XV,\\ XVII, XXII,\\ XXIII, New I,\\ New III, New V\end{tabular} & \begin{tabular}[c]{@{}c@{}}73053\\(16)\end{tabular} & \begin{tabular}[c]{@{}c@{}}57496\\(13)\end{tabular}  \\
\hline
\end{tabular}
% \end{adjustbox}
\end{adjustwidth}
\caption{Datasets and their respective train and test splits used to explore generalization. Each dataset is classified into two broad categories called \textit{outdoors} and \textit{constrained}. Eye images in outdoor datasets exhibit large proportion of environmental reflections. Constrained datasets are acquired from experiments or synthetically rendered within indoor, lab environments with little to no reflective artifacts. $^{*}$ Approximately a third of LPW recordings were collected outdoors. Please see Supplementary Table 1 for details about the splitting process.}
\label{tab:dataset_avail}
\end{table}
 
 \subsection{Network architecture}

Drift-free~\cite{Santini2019} and parallax-free~\cite{gibaldi2021solving}  eyetracking requires modelling the approximate 3D center of rotation of an eyeball from 2D pupil~\cite{Kassner2014Pupil:Interaction,Dierkes2018ARefraction,Dierkes2019APrediction} or limbus ellipses~\cite{wood2014eyetab}. This is achieved by segmenting an eye image into relevant categories and fitting ellipses to the pupil, iris and sclera boundaries. Convolutional encoder-decoder architectures have successfully been deployed to segment an eye image and extract pupil and iris ellipses~\cite{Chaudhary2019b,kothari2021ellseg,Yiu2019DeepVOG:Learning,Vera-Olmos2018DeepEye:Environments,Garbin2019OpenEDS:Dataset,Fuhl2019500000Segmentation} for datasets with pixel-level semantic annotations. Most publicly available datasets do not provide access to pixel-level ground truth annotation but instead provide the pupil center only. To circumvent this limitation, we adopt with minimal changes the EllSeg framework~\cite{kothari2021ellseg} that uniquely allows us to train a network using datasets with partial annotations. This is primarily achieved by tasking a convolutional network to segment entire elliptical masks instead of visible eye parts. The center of mass of the predicted elliptical maps allows us to optimize the entire architecture using only pupil and/or iris center annotations, conveniently allowing us to train a network on eye images with partial annotations~\cite{kothari2021ellseg}. For a complete breakdown of the architecture, please see Supplementary Figure 1.

\subsection{Normalization scheme}
Normalizing the input to a convolutional layer significantly speeds up training, improves peak accuracy performance, and reduces the dependence on weight initialization and hyper-parameter searches~\cite{ioffe2015batch,santurkar2018does}. Batch normalization in particular is a widely accepted technique to reparameterize the underlying optimization problem by providing a smoother loss landscape~\cite{santurkar2018does}. It computes $\mu_k$ and $\sigma_k$ within a population sample (e.g. a \textit{batch}) and approximates the global statistic by accumulating these values as training progresses. Statistics accumulated during the training phase are fixed during model evaluation and are generally provided alongside network parameters to facilitate inference.

Batch normalization is effective if statistics of the population sample accumulated during training are approximately equal to the global statistics of features extracted from the test set. This assumption is often violated in the context of domain generalization as statistics of the images we are evaluating our network on can be significantly different as compared to statistics observed during training. To overcome this limitation, we adopt Instance Normalization~\cite{ulyanov2016instance} as a drop-in replacement for Batch Normalization and observe improvements to generalization (see Supplementary Tables 5, 6 and 7). Instance Normalization computes the mean and standard deviation on a per image basis which are used to normalize and cast image features, irrespective of their domain, to the same unit-normal distribution.

% \subsection{Model Performance Metrics}
% We report our results on three separate metrics of performance. The pupil and iris center error which is measured in pixel units, and the mean Intersection-Over-Union (mIoU) score~\cite{eelbode2020optimization,tanimoto1968elementary,PaulJaccard1912} measured as the fraction of intersection to the union of the predicted and groundtruth categories of each pixel. Note that only the ground truth pupil center annotations are available for the majority of datasets with real eye imagery with the exception of OpenEDS, which contains all annotations. Hence, our analysis primarily focuses on pupil center performance with specific observations made for datasets with annotated iris centers and segmentation masks. 

\subsection{Generalization tests}
\label{sec:generalization_tests}
The primary goal of this work is to determine if training a model using multiple datasets results in better performance when generalized to an unseen domain as opposed to selecting the best performing domain-specific model. The former follows our intuitive understanding that training a model on multiple domains expands the available distribution we draw samples from. The latter is possible when the domain we are evaluating overlaps with an existing dataset. The adopted approach draws inspiration from Koshla~\etal~\cite{Khosla2012} and proposes four tests which allow us to explore our hypotheses. Every test below will be run on each individual domain’s test set.

\subsubsection{Within-dataset}
This test is intended to measure the ceiling performance of a model for a given dataset’s test set. Evaluating a model on the same dataset it was optimized on returns the upper performance limit. Any performance exceeding this measure indicates that a distribution gap exists between the train and test samples. This could occur if the dataset is limited by insufficient variability due to a biased sampling of its data distribution, insufficient number of subjects or limited gaze positions due to constrained tasks.

\subsubsection{Cross-dataset}
This test is intended to measure the performance of a model under conditions of cross-dataset Domain Generalization using a single training dataset. Cross dataset results indicate the performance of models when evaluated on domains not utilized during training. This test allows us to quantify how dissimilar two domains are based on their cross dataset performance measures. For every available dataset, this test allows us to empirically find the closest matching dataset.

\subsubsection{All-vs-one}
This test is intended to measure the ceiling level of performance when training utilizes all available datasets, including the within-domain dataset. This scheme trains a model using combined imagery from all available datasets. Comparing the performance of a model trained on all available distributions against its equivalent within-dataset performance gives us a clue about a single model’s ability to capture information from multiple heterogeneous distributions. Deteriorated performance indicates insufficient network capacity and serves as a test to ensure that our results are not influenced by network architecture.

\subsubsection{Leave-one-out}
This test is intended to measure the performance of a model trained using multiple datasets except on a given test set that is used to evaluate the model performance. For example, leave-one-out results on the OpenEDS test set would involve training a model with all datasets except the OpenEDS training images.  Comparing leave-one-out test results with cross-dataset performance provides evidence for the optimal strategy which maximally generalizes on the OpenEDS dataset.

\subsection{Predictions}
In this section, we briefly summarize expected results using the four proposed tests. Model performance will be reported on their relative ability to estimate the pupil and iris centers in units of pixels and mean Intersection-over-Union (mIoU) scores~\cite{eelbode2020optimization,tanimoto1968elementary,PaulJaccard1912}. Systematic comparison across the proposed tests allows us to generate specific predictions directly related to our hypotheses.

\subsubsection{Hypothesis 1: Cross-domain generalization}
A single model trained on data drawn from multiple heterogeneous domains (the leave-one-out test) can generalize better than a specialized, dataset-specific model when evaluated on domains \textbf{unseen during training} (the cross-dataset test). This is achieved by comparing the leave-one-out test with the cross-dataset test. If the leave-one-out test leads to better generalization than the cross-dataset test, this would indicate that adding more datasets and acquiring a broader distribution will improve generalization, presumably asymptotically until the within-dataset performance limit is reached. If the cross-dataset model outperforms the leave-one-out test then it suggests that the best approach to achieve generalizability is to explore test-time techniques to find an optimal model from a pool of dataset-specific models, rather than to rely upon a single broadly-trained and general-purpose model.

\subsubsection{Hypothesis 2: Within-domain generalization}
Researchers often develop models for their specific applications with a fixed hardware setup and environmental conditions. This involves the collection and annotation of data we want to evaluate the model on. This is a time consuming, expensive process and researchers often collect and annotate a smaller subset with the aim of producing a model which generalizes to the entire breadth of data. However, a model may not generalize across subjects or gaze positions due to distribution shifts caused by limited or biased sampling of their specific distribution. The within-dataset performance represents the distribution shift exhibited when the train and test splits are sampled from the same distribution. We hypothesize that any remaining distribution shift due to biased sampling will be mitigated by increasing the breadth of training distribution. If this is true, then the all-vs-one test will outperform the within-dataset test. If the within-dataset test outperforms the all-vs-one test then it suggests that the all-vs-one model has insufficient complexity (parameters) to capture the entire breadth of the training data. Lastly, we can conclude that a dataset has been optimally sampled if both both tests exhibit nearly the same performance. All possibilities are explored in post-hoc tests and addressed in the discussion section.

\subsubsection{Hypothesis 3: Effects of data augmentation}
Data augmentation has numerous benefits for machine learning applications~\cite{hernandez2018further,nalepa2019data}. It improves generalization by combating overfitting, expands the training distribution and reduces domain gap by combating prior shifts. Previous work has demonstrated that domain-specific data augmentation significantly improves cross-subject performance in context of head-mounted eyetracking~\cite{Eivazi2019ImprovingAugmentation,Park2018DeepEstimation}. Data augmentation is generally accepted as standard practice within Machine Learning and we quantify its effects on generalization.

It is plausible that within the context of eye image segmentation, data augmentation sufficiently reduces distribution shift by expanding the available training distribution while removing the need for multiset training. If this is true, then we expect cross-dataset tests to outperform the leave-one-out test. If this is false, then the inclusion of data augmentation is expected to improve performance across all tests while not affecting relative performance. Table~\ref{tab:chapter_7:aug_schemes} summarizes all augmentation schemes explored in this manuscript.

\subsection{Analysis}
Exploring our hypotheses for each dataset requires us to compare model performance between the proposed tests while keeping the within-dataset performance as baseline. However, datasets differ in complexity. A 1 pixel error improvement in pupil center estimation on a particular dataset cannot be compared objectively with others without considering the variability inherent within the ground truth, for the reason that one would expect more variability in model estimates for an inherently more variable test dataset. To mitigate this limitation, we consider the dispersion of test results (reported in median \rev{absolute} deviation units, or MADs, due to its robustness to outliers) which are derived by normalizing a performance metric by the within-dataset test result.

\subsection{Training details}
This work relies on a modified version of DenseElNet~\cite{kothari2021ellseg}, a standard encoder-decoder convolutional neural network inspired by RITNet~\cite{Chaudhary2019b}, FC-DenseNet~\cite{Jegou2017TheSegmentation} and UNet~\cite{Ronneberger2015U-net:Segmentation}. Eye images are passed to the encoder which consists of 4 densely connected convolutional blocks. Each block extracts features from eye images while down-sampling their spatial extent by a factor of 2. Latent representations rich with semantic features are then fed into the decoder which produces a segmentation output mask for each eye image. For a complete breakdown of architecture, please refer to Supplementary Figure 1. Latent representations are also fed into a regression module which regresses pupil and iris ellipse parameters, if available.

For experiments which involve joint optimization on multiple datasets, an equal number of randomly selected samples from each domain are concatenated as input to our neural network. This process alleviates the concern of unequal representation due to a disproportionate number of samples in each domain (see Table~\ref{tab:dataset_avail}). The validation set comprises 20\% samples held out from the training set, either from a single or multiple domains. All models are optimized with ADAM~\cite{Kingma2014Adam:Optimization} for 80 epochs. This work utilizes the loss functions proposed in EllSeg (Section~3.3 in Kothari~\etal~\cite{kothari2021ellseg}). To curb overfitting and to ensure our model can be applied to both, left and right eyes, each eye image and its associated annotations are randomly flipped horizontally.

Experiments involving data augmentation modify input eye images and its associated annotations by randomly selecting an operation from a pool of augmentation schemes. Table~\ref{tab:chapter_7:aug_schemes} provides a summary of all schemes explored in this manuscript. \rev{Along with the ten data augmentation schemes introduced in Table~\ref{tab:chapter_7:aug_schemes}, we also incorporate the possibility of not altering an eye image. Hence for any given eye image sample, we select one out of eleven possible choices, \ie each augmentation scheme has a $\nicefrac{1}{11}$ chance of being selected.}

\begin{table}[]
\centering
\renewcommand{\arraystretch}{1.25}
\resizebox{0.8\textwidth}{!}{%
\begin{tabular}{|c|c|c|c|c|}
\hline
Gauss blur &
  Motion blur &
  Gamma &
  Exposure &
  Gauss noise \\ \hline
\begin{tabular}[c]{@{}c@{}}w=7\\ $\sigma$=$\mathcal{U}$(2,7)\end{tabular} &
  \begin{tabular}[c]{@{}c@{}}w=7\\ $\theta$=$\mathcal{U}$(0, $\pi$)\end{tabular} &
  $\gamma$=$\mathcal{U}$(0.6, 1.4) &
  \begin{tabular}[c]{@{}c@{}}$\Delta$L$^i_{max}$=0.8 $\times$ $\tilde{L}^i$\\ $\Delta$L$^i_{min}$=0.8 $\times$ (255 - $\tilde{L}^i$)\\ L = L + $\mathcal{U}$(-$\Delta$L$^i_{min}$,$\Delta$L$^i_{max}$)  \end{tabular} &
  \begin{tabular}[c]{@{}c@{}} $\mu$=0 \\ $\sigma$=$\mathcal{U}$(2, 16) \end{tabular} \\ \hline
Synth lines &
  Scale &
  Rotation &
  Translation &
  Synth fog \\ \hline
\begin{tabular}[c]{@{}c@{}}$\mathcal{U}$(1,10) \\ random\\ white lines\end{tabular} &
  \begin{tabular}[c]{@{}c@{}}Scale\\ factor \\ $\mathcal{U}$(0.5, 0.9) \end{tabular} &
  $\theta$=$\mathcal{U}$(-$\pi$,$\pi$)/4 &
  \begin{tabular}[c]{@{}c@{}}Horz: {$\mathcal{U}$(-W, W)/3}\\ Vert: {$\mathcal{U}$(-H, H)/3} \end{tabular} &
  \begin{tabular}[c]{@{}c@{}c@{}}Please refer to\\ \textit{imgaug}\\~\cite{imgaug}\end{tabular} \\ \hline
\end{tabular}
}
\vspace{0.25cm}
% \captionsetup{singlelinecheck=off, margin={0.5cm, 0.5cm}, format=hang}
\caption{Augmentation schemes applied to every single eye image with a $\nicefrac{1}{11}$ probability. The probability of 1/11 reflects that there are 10 possible augmentations in addition to the lack of augmentation. $\mathcal{N}$ and $\mathcal{U}$ indicate a normal and uniform distribution respectively. The median iris intensity $\mathrm{\hat{L}}$, if available, or a range of $\pm$50 intensity levels is utilized during image exposure augmentation.}
\label{tab:chapter_7:aug_schemes}
\end{table}

\subsection{Evaluation criterion}

The performance of a model on a dataset is evaluated on a set of eye images (test set) from human subjects that were never present during training (see Table~\ref{tab:dataset_avail}). Twenty percent of training data is withheld for model validation during training. The best performing model is selected as the configuration which maximizes an average of mIoU and $d_i$ + $d_p$ on the validation set. Here, $d_p = 1-\alpha e_p$ and $d_i=1-\alpha e_p$ where $e_p$ and $e_i$ are pixel errors in predicting the pupil and iris centers and $\alpha=\nicefrac{1}{240}$ is derived from the smallest image dimension. All models are evaluated every 2000 \textit{iterations}, wherein an iteration is defined as a single network parameter update operation based on a batch of eye images. Annotations not present in a dataset are ignored while computing the evaluation metric. For experiments which involve training on multiple datasets, each batch consists of 3 eye images extracted randomly from every dataset included in the experiment. For experiments involving training on a single dataset, each batch consists of 24 eye images.

\begin{figure}
\centering
%\begin{adjustwidth}{3cm}{0cm}
\begin{subfigure}{.48\textwidth}
  \centering
  \captionsetup{singlelinecheck=off, margin={0.5cm, 0.5cm}, format=hang}
  \includegraphics[width=\linewidth]{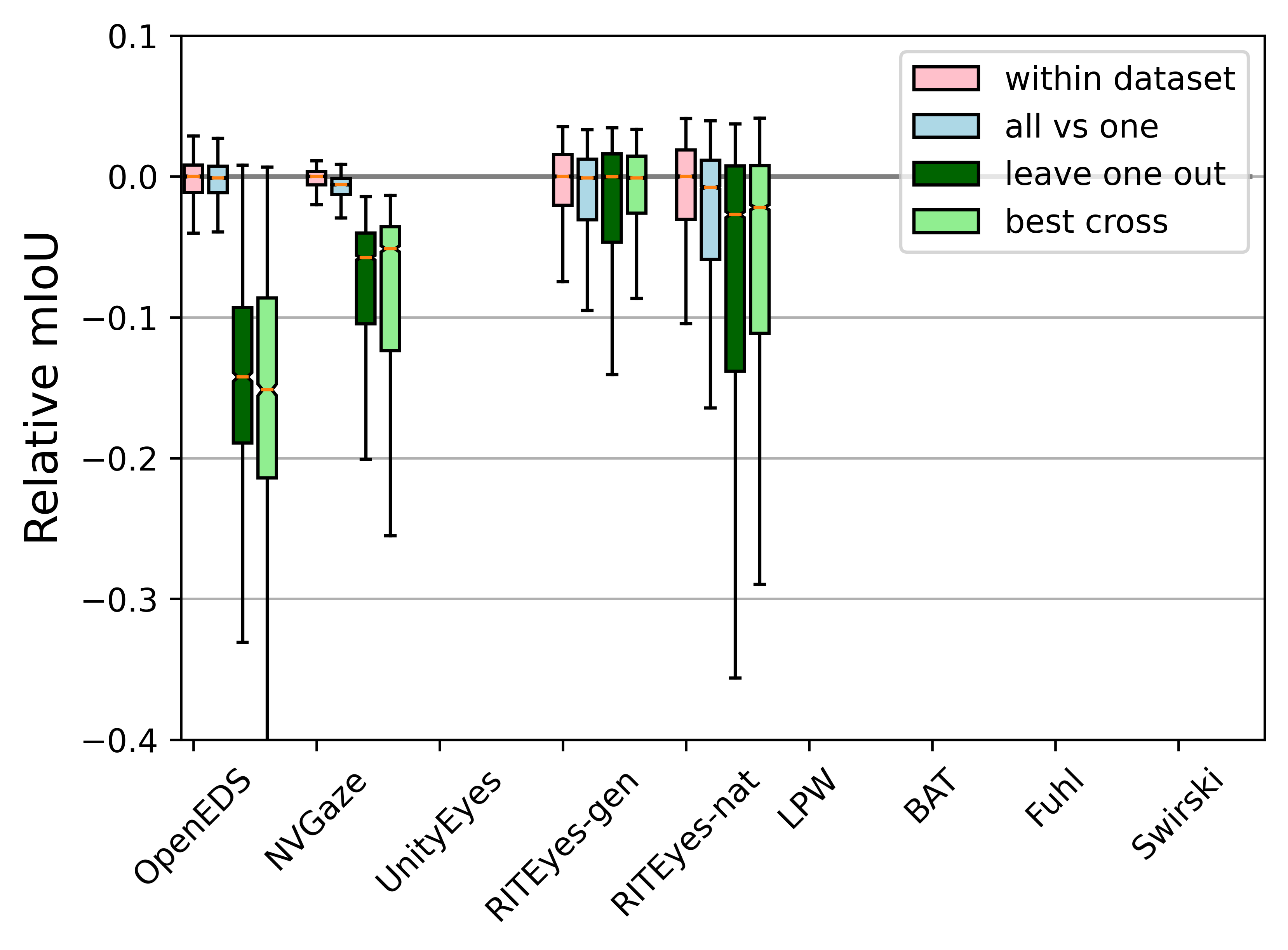} 
  \caption{Performance measured as the mIoU metric. The Y axis represents IoU score relative to the control condition, \textit{within-dataset}. Positive values indicate a performance improvement.}
\end{subfigure}
\begin{subfigure}{.48\textwidth}
  \centering
  \captionsetup{singlelinecheck=off, margin={0.5cm, 0.5cm}, format=hang}
  \includegraphics[width=\linewidth]{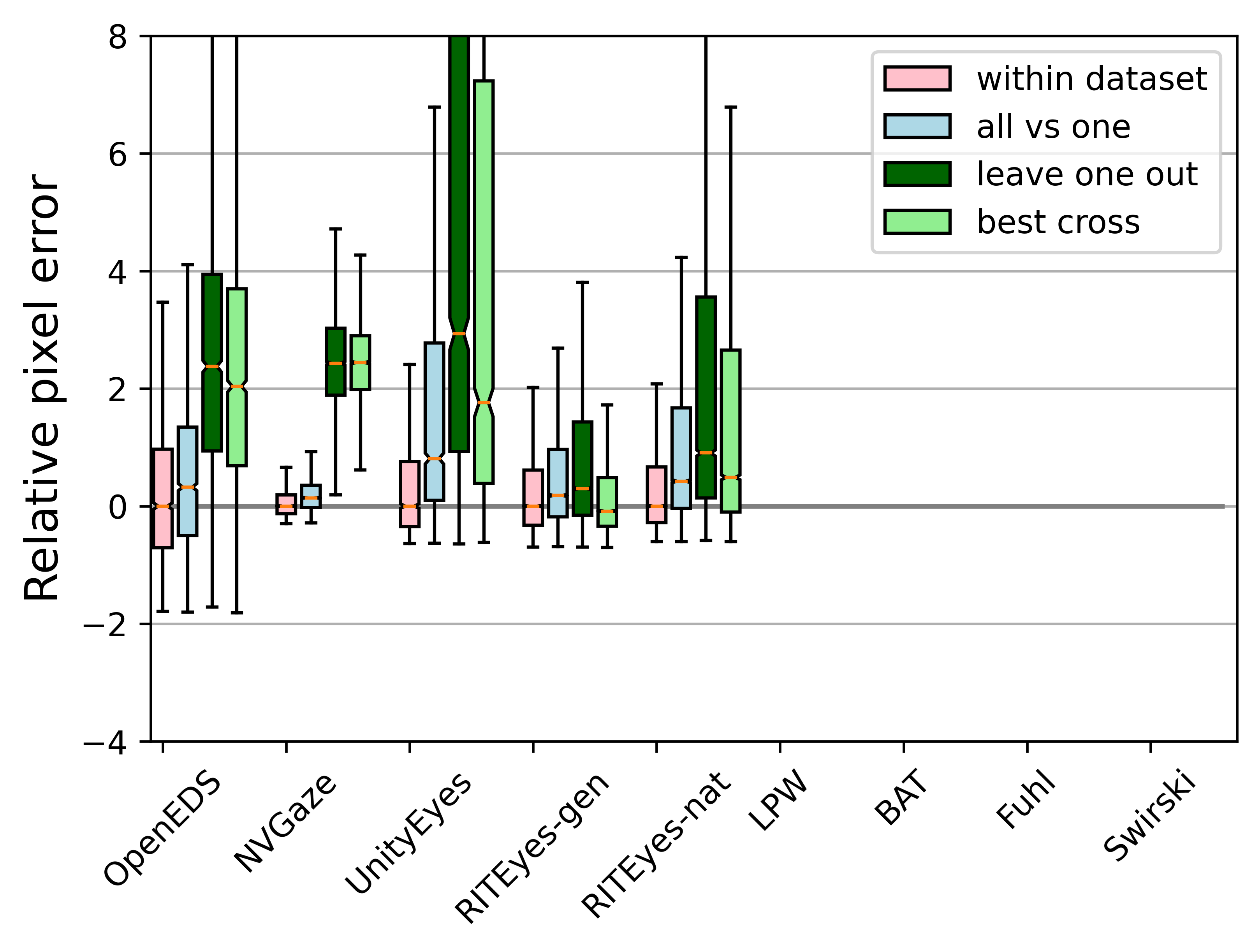}  
  \caption{Error in iris center, $e_i$, in pixels. The Y axis represents pixel distance relative to the control condition, \textit{within-dataset}. Negative values indicate a performance improvement.}
\end{subfigure}

\begin{subfigure}{.8\textwidth}
  \centering
  \includegraphics[width=0.9\linewidth]{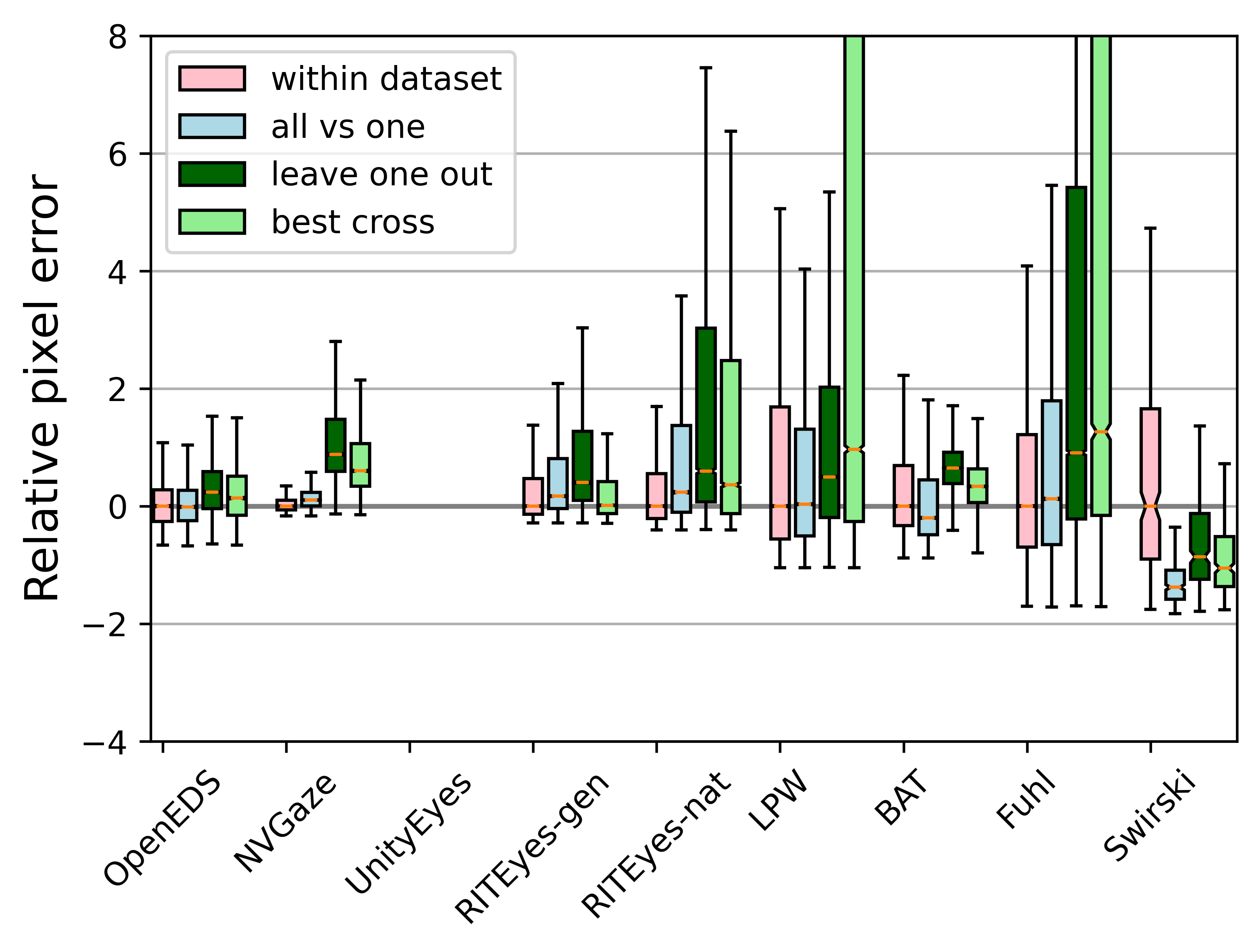}
  \caption{Error in pupil center, $e_p$, in pixels. The Y axis represents pixel distance relative to the control condition, \textit{within-dataset}.Negative values indicate a performance improvement.}
\end{subfigure}
%\end{adjustwidth}
\caption{Generalization test results. Each box plot highlights a model’s performance centered to the within-dataset limit for each domain. The line and notch present within each box plot represents the median and 95\% confidence interval respectively while the ends of each box denotes the 1$^\mathrm{st}$ and 3$^\mathrm{rd}$ quartile. All images are 320$\times$240 resolution. Note that datasets which are missing groundtruth annotations do not have a boxplot entry. All measures are reported relative to the within-dataset performance, which necessarily lies at zero. The absolute performance of various tests are provided in Supplementary Tables 2, 3 and 4}
\label{fig:gen_results_aug_0}
\end{figure}

\subsection{Hypothesis 1: Training with multiple datasets is an optimal strategy for generalization.}
\label{sec:chapter_7:hypothesis_1}

\section{Results and Discussion}
In this section, we provide results for tests listed in Section~\ref{sec:generalization_tests}. We focus our observations primarily on datasets with real eye images since observations on synthetic datasets do not support evidence for generalization during naturalistic conditions. \rev{We report our results on three separate metrics of performance. The pupil and iris center error which is measured in pixel units, and the mIoU score measured as the fraction of intersection to the union of the predicted and groundtruth categories of each pixel. Note that only the ground truth pupil center annotations are available for the majority of datasets with real eye imagery with the exception of OpenEDS, which contains all annotations. Hence, our analysis primarily focuses on pupil center performance with specific observations made for datasets with annotated iris centers and segmentation masks}. 

Figure~\ref{fig:gen_results_aug_0} reveals that a multiset model offers improvements to generalization as opposed to selecting any dataset specific model for the LPW and Fuhl datasets (LPW $\uparrow$0.66 MADs, Fuhl $\uparrow$0.42 MADs). This is evident when comparing the best performing cross-dataset test result with the leave-one-out test. However, results indicate the opposite to be true for OpenEDS, BAT and Swirski datasets ($\downarrow$0.37, $\downarrow$0.71 and $\downarrow$0.18 respectively). One explanation for this behavior is the conditions which these dataset represent. The Fuhl datasets were collected from subjects tasked to drive a car outdoors while subjects in LPW were tasked to fixate on a red ball moved by an experimenter in a variety of environmental conditions. In contrast, OpenEDS, BAT and Swirski datasets consist of physically restrained subjects who were given  tasks which elicited a narrow range of gaze positions. Furthermore, eye images in these datasets were collected under constrained conditions with fixed lighting. This observation alludes to the variability of eye images inherent in these datasets.

To further analyze the dimensions along which these datasets vary, we performed a post-hoc analysis on the pupil location and appearance across eye images in every dataset present (see Supplementary Figure 2). We observe that eye images in the LPW and Fuhl datasets exhibit a broader distribution of pupil positions. They also demonstrate higher variability in the proportion of iris pixels across eye images. Humans exhibit relatively constant iris size when compared to the pupil, which frequently constricts and dilates. A large variation in proportion of iris pixels indicates a wider range of eye camera orientation and position. This is verified by Tonsen~\etal who frequently varied the eye camera pose between frontal to off-axis views to capture a broader appearance of eye images~\cite{tonsen2016labelled}. Furthermore, the luminance distribution of pupils in these two datasets demonstrate a large spread and significant presence of intensities above the mean luminance of an eye image. Note that due to the absence of groundtruth, the luminance distribution of pupil regions is derived from predicted segmentation masks in Supplementary Figure 2.

The multiset training paradigm also demonstrates an improvement in segmentation performance on the OpenEDS ($\uparrow$0.95 MADs). This evidence supports the intuition that multiset training offers a broader range of iris appearances which in turn has the potential to improve iris segmentation. To summarize, our analysis supports the conclusion that a multiset training paradigm offers better generalization as opposed to selecting the best performing cross-dataset model for eye images acquired under conditions with high variability.

\subsection{Hypothesis 2: Training with multiple datasets will improve within-dataset performance.}
\label{sec:chapter_7:hypothesis_2}
The within-dataset performance represents the distribution shift exhibited when the train and test splits are sampled from the same distribution. We hypothesize that training with multiple heterogeneous distributions could improve upon the within-dataset performance by mitigating this latent distribution shift. That is to say we hypothesize that in the presence of distribution shift, the shift will be mitigated by increasing the breadth of the training distribution, as would be suggested if the all-vs-one test were to outperform the within dataset test. However, results indicate that it is only on the BAT and Swirski datasets that the all-vs-one test outperforms the within-dataset performance, by 0.45 and 1.34 MADs, respectively. For the majority of the datasets, the all-vs-one model demonstrates only a slight reduction in performance from the expected within-dataset upper bound. The highest disagreement with the upper bound is evidenced on the Fuhl dataset with $\downarrow$0.146, on the pupil center upper bound respectively.

Closer inspection reveals that the Swirski dataset is comprised of $\sim$300 eye images from a single subject and that the BAT dataset consists of 3.5K eye images from 3 subjects. This is evidenced by the sparse pupil center distribution and the poor overlap exhibited by the train and test eye images (see Supplemental Figure 2). Results indicate that these datasets are insufficient for cross-subject generalization despite their images being drawn from the same environment and eyetracker hardware. This limitation is mitigated by the all-vs-one model which provides a broad training distribution and improving generalization to new subjects.

\subsection{Hypothesis 3: Effects of data augmentation for generalization}

\begin{figure}
\centering
\begin{subfigure}{.48\textwidth}
  \centering
  \captionsetup{singlelinecheck=off, margin={0.5cm, 0.5cm}, format=hang}
  \includegraphics[width=\linewidth]{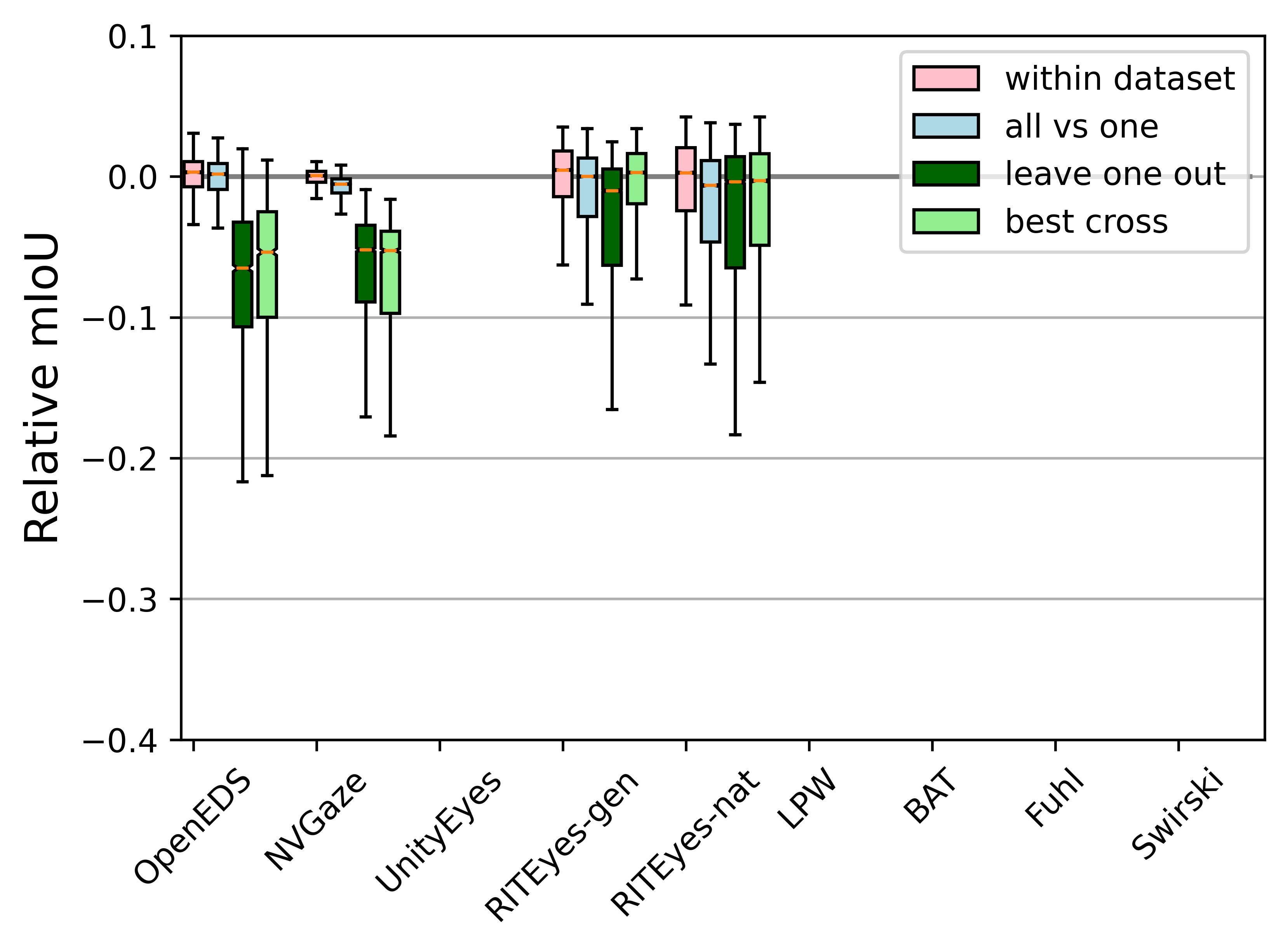}  
  \caption{Performance following data augmentation measured as the mIoU metric relative to training in the absence of augmentation. Positive values indicate an improvement in performance.}
\end{subfigure}
\begin{subfigure}{.48\textwidth}
  \centering
  \captionsetup{singlelinecheck=off, margin={0.5cm, 0.5cm}, format=hang}
  \includegraphics[width=\linewidth]{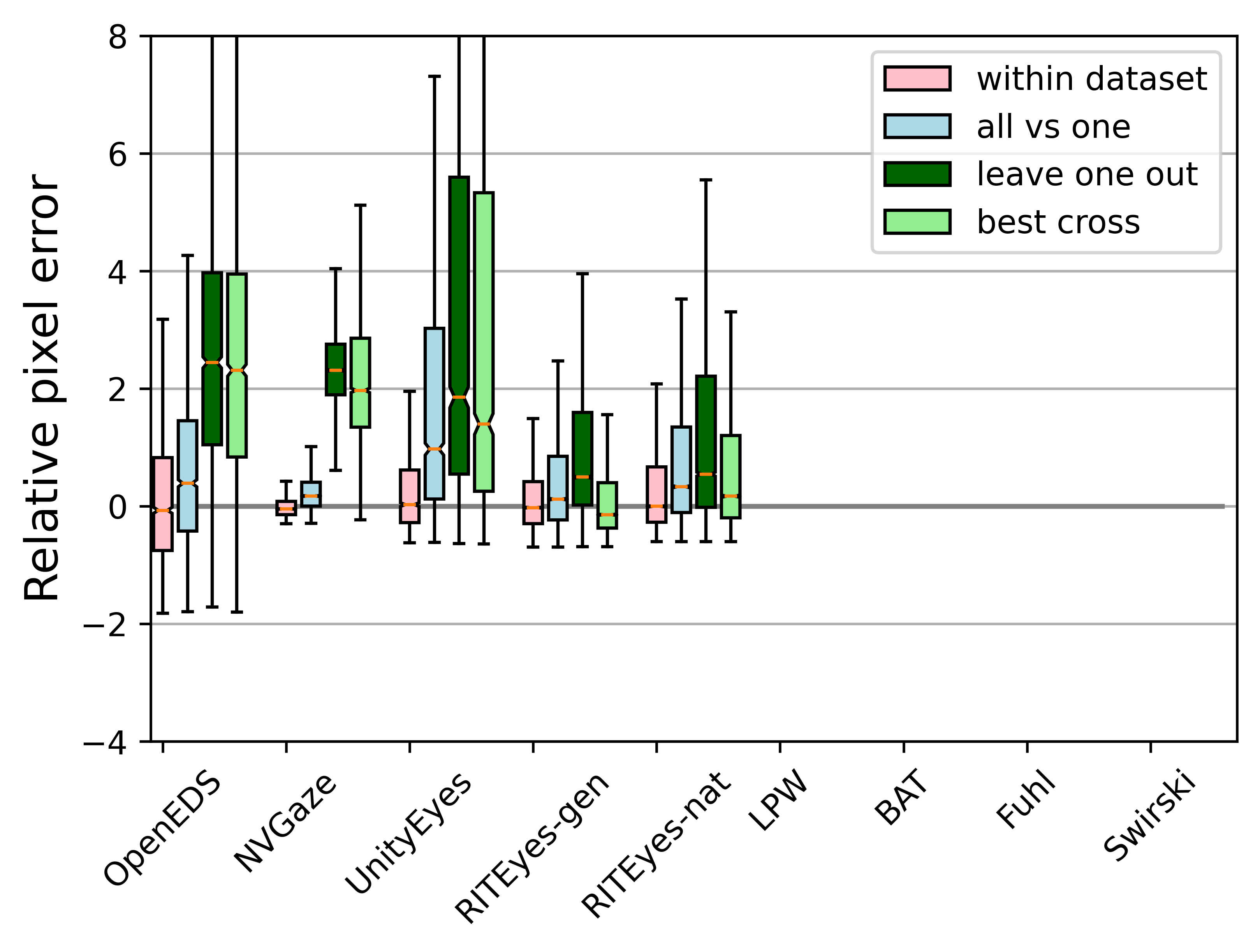}
  \caption{Performance following data augmentation measured as error in iris center, $e_i$, in pixels. Negative values indicate an improvement in performance.}
\end{subfigure}

\begin{subfigure}{.8\textwidth}
  \centering
  \includegraphics[width=0.9\linewidth]{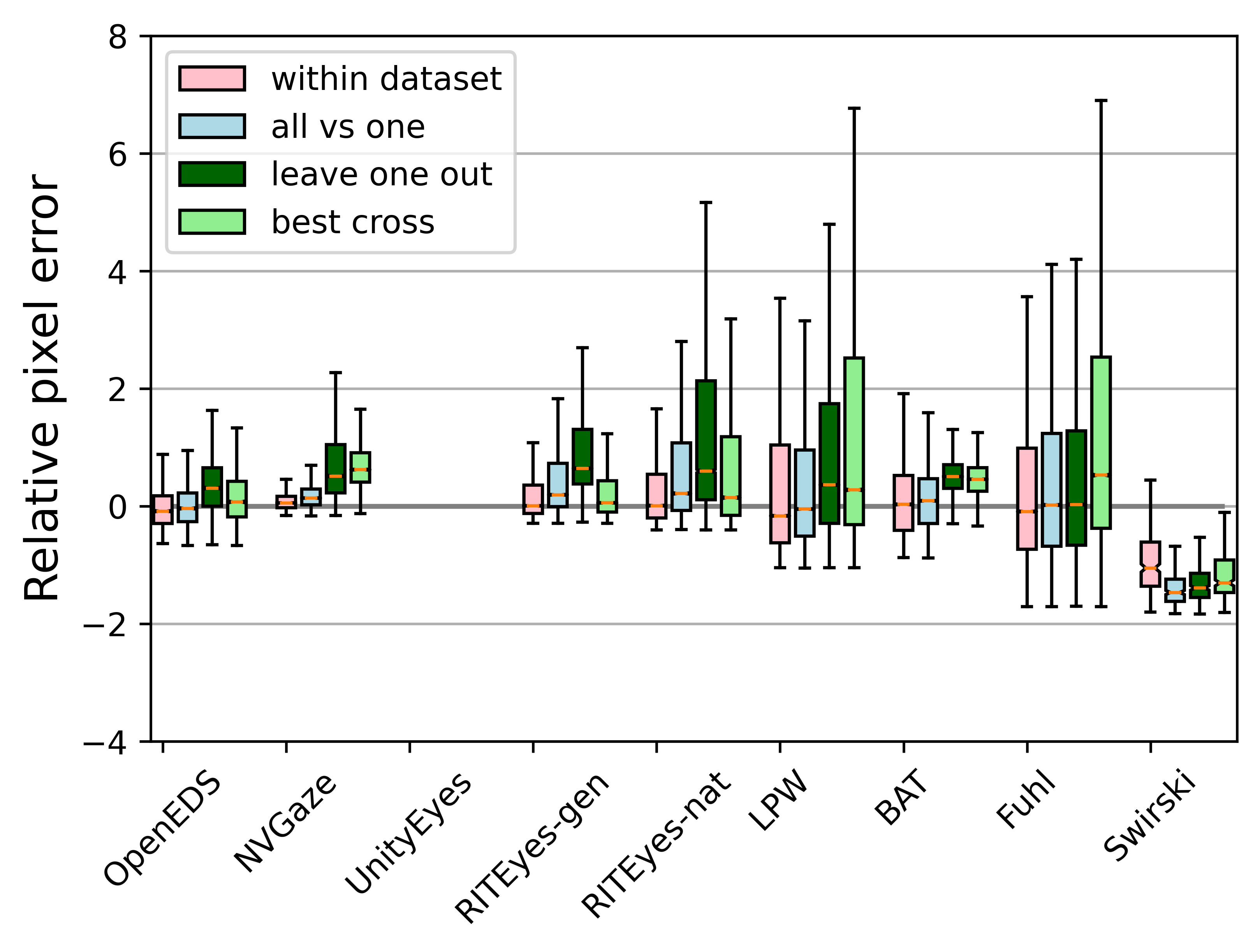}
  \caption{Performance following data augmentation measured as error in pupil center $e_p$ in pixels. Negative values indicate an improvement in performance.}
\end{subfigure}
\caption{Generalization test results to study the effects of data augmentation. Each box plot highlights a model’s performance centered to the within-dataset limit for each domain. The line and notch present within each box plot represents the median and confidence interval respectively while the ends of each box denotes the 1$^\mathrm{st}$ and 3$^\mathrm{rd}$ quartile. All images are 320$\times$240 resolution. Note that boxplots pertaining to datasets without a certain groundtruth annotation are missing. All measures are centered to the within-dataset threshold without data augmentation as reported in Figure 2 and Supplementary Tables 2, 3 and 4}
\label{fig:gen_results_aug_1}
\end{figure}

Figure~\ref{fig:gen_results_aug_1} summarizes performance gains observed by augmenting eye images, and reveals that data augmentation improves performance across all testing sets. This is indicated by reductions in the inter-quartile range and improvements to the median performance across all metrics and tests. 

 We find that our previous established hypotheses (see Section~\ref{sec:chapter_7:hypothesis_2}) regarding multiset training remains unaffected by the presence of data augmentation. Significant improvements to generalization can be observed on the Fuhl datasets wherein a combination of multiset training and data augmentation achieves performance on par with the within-dataset baseline.
 
\begin{figure}
     \centering
     \includegraphics[width=1.00\linewidth]{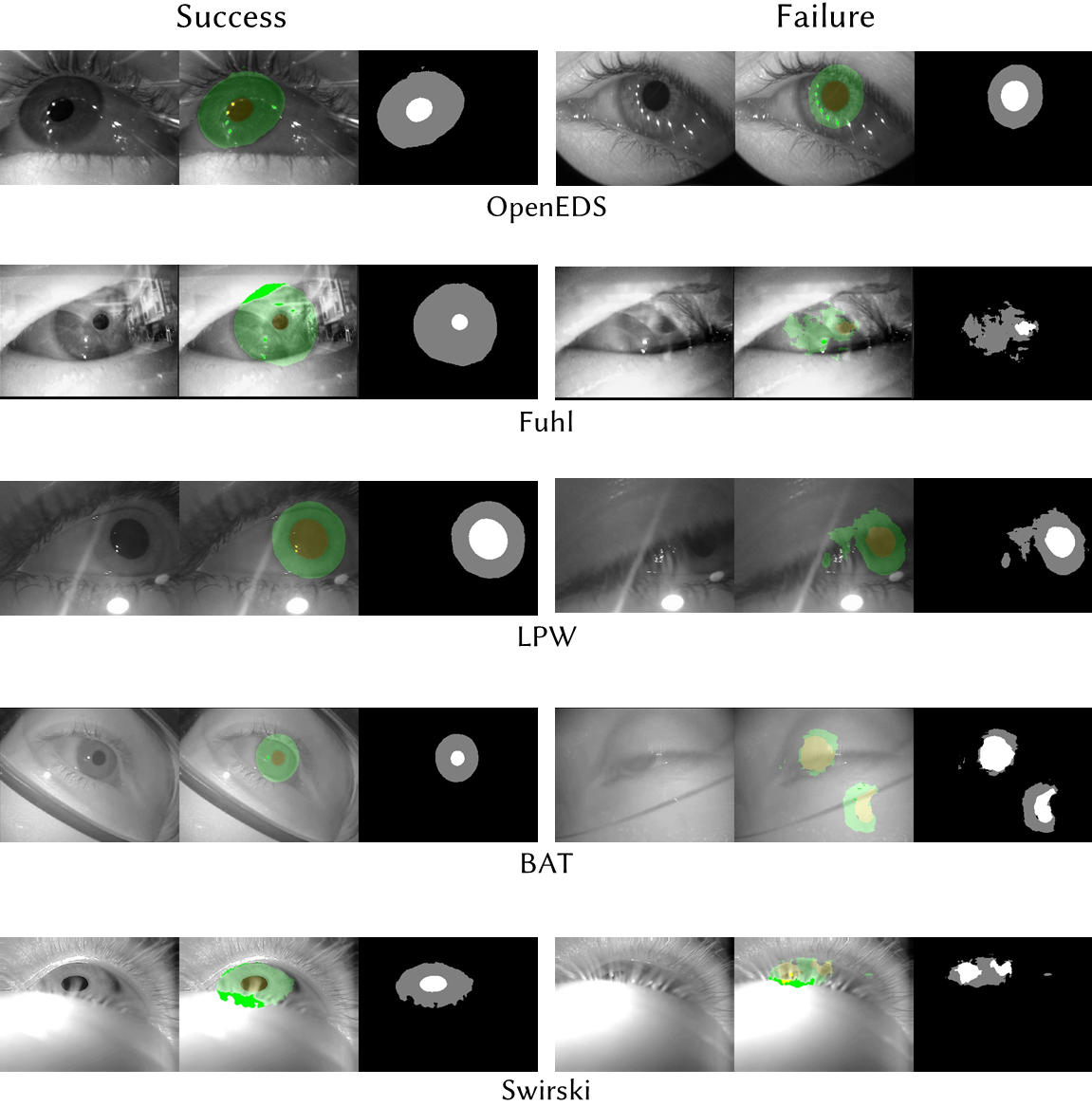}
     \caption{Success and failure examples of DenseElNet predictions in the leave-one-out configuration. Note that for any eye image shown in this figure, not a single image from its domain/dataset was used to train the model. This figure is representative of what to expect during model deployment on the datasets shown.}
     \label{fig:leave_one_out_op_quality}
 \end{figure}

 Our tests of hypothesis 1 revealed that the LPW dataset favors multiset training, possibly due to its relatively wide distribution of pupil center locations. As expected, data augmentation sufficiently reduces the performance gap between the multiset model and the best performing cross-dataset model (albeit with a higher spread), indicating that either approach for generalization returns similar performance. Please see Figure~\ref{fig:leave_one_out_op_quality} to visualize the performance of a multiset model in leave-one-out configuration.
 
 In hypothesis 2, we identified that multiset training aids in overcoming the effects of limited sampling. Previously, the BAT all-vs-one test outperformed the within-dataset baseline indicating a distribution shift between its own train and test set. This shift is entirely eliminated by artificially augmenting eye images. The multiset training however outperforms the within-dataset baseline for the Swirski dataset, indicating that data augmentation alone does not sufficiently expand the training distribution for datasets with significant limited sampling.
 
 \subsection{Effects of reducing model complexity}
 Head-mounted eyetracking is primarily used for studying eye movements during unconstrained activities. This requires algorithms or models to be computationally inexpensive in order to conserve battery or reduce system latency to support downstream applications which rely on gaze estimates. The model employed in this work contains 2.24 million trainable parameters and operates at 118Hz on a NVIDIA RTX3090. When the number of trainable parameters was reduced to 529K, performance speed increased to 140Hz. Although this increase in speed is accompanied by an overall reduction in model performance across all tests, the ordinal relationships between tests remain unaffected by model complexity (see Supplementary Figure 3). 
 
\section{Conclusion}
In this work, we have explored the task of Domain Generalization for segmenting near infrared eye images. This was achieved by jointly optimizing an encoder-decoder network on multiple eye image datasets with the intuition that a model would learn a generalized representation of eye images and elliptical eye parts. This work evaluated two approaches towards generalization, a) a multiset training approach to produce a single, robust model or b) pick the best performing model from a pool of pretrained, dataset-specific models. \rev{Generalization results indicate that datasets which exhibit higher variability benefit from multiset optimization.} In contrast, dataset-specific models generalize better onto constrained use-cases, such as eye tracking applications where the eye camera properties are fixed, camera pose is constrained, and lighting conditions are controlled.

Although a model's peak performance can be attained when it is trained and evaluated on data sampled from the same source distribution, limited or biased sampling often leads to a distribution shift and deteriorated model performance. Results indicate that multiset training can be utilized to identify and mitigate such a distribution shift, if it exists. This work also validates the benefits of data augmentation to further improve generalization by observing improvements in model performance across the board. Results indicate that a combination of data augmentation and multiset training attains the peak performance recorded for unconstrained eye images. While data augmentation does improve performance, it does not invalidate the contributions of multiset learning but instead, complements it. Models and code to facilitate our experiments are publicly available~\footnote{\href{https://bitbucket.org/RSKothari/multiset_gaze/src/master}{bitbucket.org/RSKothari/multiset\textunderscore gaze/src}} for the research community.

\begin{acks}

% Acknowledgments withheld for review.
The authors thank Research Computing at the Rochester Institute of Technology for their support and providing necessary hardware. The authors would also like to thank Dr. Aneesh Rangnekar for their helpful comments and feedback.
\end{acks}


\newpage
\FloatBarrier
\bibliographystyle{ACM-Reference-Format}
\bibliography{main}
\FloatBarrier
\vspace*{1cm}
Received November 2021; revised January 2022; accepted April 2022

\newpage
\FloatBarrier

\setcounter{table}{0}
\setcounter{figure}{0}

\section*{Supplementary data}
\FloatBarrier
% \usepackage{booktabs}
\begin{table}[]
\renewcommand{\arraystretch}{1.2}
\begingroup
\begin{tabular}{l|l|r|r} 
\toprule
\textbf{Dataset} & \textbf{Splitting procedure}                                                                                                                                                                                                                                                                                                           & \multicolumn{1}{l|}{\begin{tabular}[c]{@{}l@{}}\textbf{Test to}\\\textbf{ train ratio}\\\textbf{ of subjects}\end{tabular}} & \multicolumn{1}{l}{\begin{tabular}[c]{@{}l@{}}\textbf{Test to}\\\textbf{ train ratio}\\\textbf{ of images}\end{tabular}}  \\ 
\hline
S-General        & Maintain splits adopted in Nair~~                                                                                                                                                                                                                                                                                                      & 0.33                                                                                                                        & 0.3                                                                                                                       \\ 
\hline
OpenEDS'19       & Adopted splits proposed by Garbin~~                                                                                                                                                                                                                                                                                                    & 0.29                                                                                                                        & 0.27                                                                                                                      \\ 
\hline
NVGaze           & Random selection                                                                                                                                                                                                                                                                                                                       & 0.25                                                                                                                        & 0.25                                                                                                                      \\ 
\hline
LPW              & \begin{tabular}[c]{@{}l@{}}Previously reported pupil detection\\rates by Fuhl~was utilized to categorize\\each subset as ~\textit{difficult} or \textit{easy}. Each\\subset was assigned to the train and test\\split such that the number of\\ \textit{difficult} and \textit{easy} subsets remained \\the same in both splits.\end{tabular} & 0.83                                                                                                                        & 0.73                                                                                                                      \\ 
\hline
Swirski          & \begin{tabular}[c]{@{}l@{}}We chose to place subject 2 in the test split\\due to the eye camera positioned in an~off-\\axis configuration\end{tabular}                                                                                                                                                                                 & 1                                                                                                                           & \multicolumn{1}{l}{}                                                                                                      \\ 
\hline
BAT              & \begin{tabular}[c]{@{}l@{}}Subjects 1 and 4 wore glasses which was\\presumed to represent difficult conditions\\ for our eye image segmentation model.\\We ensured difficult conditions were present\\ in both train and test splits by assigning subject\\1 to the train spit and subject 4 to the~test split.\end{tabular}           & 1                                                                                                                           & 0.96                                                                                                                      \\ 
\hline
S-Natural        & Maintain splits adopted in Nair~~                                                                                                                                                                                                                                                                                                      & 0.33                                                                                                                        & 0.33                                                                                                                      \\ 
\hline
UnityEyes        & Random selection                                                                                                                                                                                                                                                                                                                       & \multicolumn{1}{l|}{-}                                                                                                      & 0.12                                                                                                                      \\ 
\hline
Fuhl             & \begin{tabular}[c]{@{}l@{}}Previously reported pupil detection rates by\\Fuhl was~utilized to categorize each subset\\as a \textit{difficult} or \textit{easy}. Each subset was~\\assigned to the train and test split such that\\the overall number of \textit{difficult} and \textit{easy}\\ subsets remained the same.\end{tabular} & 0.81                                                                                                                        & 0.78                                                                                                                      \\
\bottomrule
\end{tabular}
\caption{Splitting criterion employed while generating individual train and test splits for each dataset}
\endgroup
\label{tab:splitting_criterion}
\end{table}
\begin{table}[]
\centering
\begin{adjustbox}{angle=90}
\begin{tabular}{cccccccccc|}
\hline
\multicolumn{1}{|c|}{Train on / Test on} &
  \multicolumn{1}{c|}{OpenEDS} &
  \multicolumn{1}{c|}{NVGaze} &
  \multicolumn{1}{c|}{UnityEyes} &
  \multicolumn{1}{c|}{RITEyes-Gen} &
  \multicolumn{1}{c|}{RITEyes-Nat} &
  \multicolumn{1}{c|}{LPW} &
  \multicolumn{1}{c|}{BAT} &
  \multicolumn{1}{c|}{Fuhl} &
  Swirski \\ \hline
\multicolumn{1}{|c|}{OpenEDS} &
  \multicolumn{1}{c|}{0.957} &
  \multicolumn{1}{c|}{0.670} &
  \multicolumn{1}{c|}{} &
  \multicolumn{1}{c|}{0.576} &
  \multicolumn{1}{c|}{0.419} &
   &
   &
   &
   \\ \cline{1-3} \cline{5-6}
\multicolumn{1}{|c|}{NVGaze} &
  \multicolumn{1}{c|}{0.652} &
  \multicolumn{1}{c|}{0.988} &
  \multicolumn{1}{c|}{} &
  \multicolumn{1}{c|}{0.563} &
  \multicolumn{1}{c|}{0.362} &
   &
   &
   &
   \\ \cline{1-3} \cline{5-6}
\multicolumn{1}{|c|}{UnityEyes} &
  \multicolumn{1}{c|}{} &
  \multicolumn{1}{c|}{} &
  \multicolumn{1}{c|}{} &
  \multicolumn{1}{c|}{} &
  \multicolumn{1}{c|}{} &
   &
   &
   &
   \\ \cline{1-3} \cline{5-6}
\multicolumn{1}{|c|}{RITEyes-Gen} &
  \multicolumn{1}{c|}{0.796} &
  \multicolumn{1}{c|}{0.936} &
  \multicolumn{1}{c|}{} &
  \multicolumn{1}{c|}{0.962} &
  \multicolumn{1}{c|}{0.933} &
   &
   &
   &
   \\ \cline{1-3} \cline{5-6}
\multicolumn{1}{|c|}{RITEyes-Nat} &
  \multicolumn{1}{c|}{0.806} &
  \multicolumn{1}{c|}{0.932} &
  \multicolumn{1}{c|}{} &
  \multicolumn{1}{c|}{0.961} &
  \multicolumn{1}{c|}{0.955} &
   &
   &
   &
   \\ \cline{1-3} \cline{5-6}
\multicolumn{1}{|c|}{LPW} &
   &
   &
   &
   &
  \multicolumn{1}{c|}{} &
   &
   &
   &
   \\ \cline{1-1}
\multicolumn{1}{|c|}{Santini} &
   &
   &
   &
   &
  \multicolumn{1}{c|}{} &
   &
   &
   &
   \\ \cline{1-1}
\multicolumn{1}{|c|}{Fuhl} &
   &
   &
   &
   &
  \multicolumn{1}{c|}{} &
   &
   &
   &
   \\ \cline{1-1}
\multicolumn{1}{|c|}{Swirski} &
   &
   &
   &
   &
  \multicolumn{1}{c|}{} &
   &
   &
   &
   \\ \cline{1-6}
 &
   &
   &
   &
   &
   &
   &
   &
   &
   \\ \cline{1-3} \cline{5-6}
\multicolumn{1}{|c|}{all-OpenEDS} &
  \multicolumn{1}{c|}{0.815} &
  \multicolumn{1}{c|}{0.985} &
  \multicolumn{1}{c|}{} &
  \multicolumn{1}{c|}{0.977} &
  \multicolumn{1}{c|}{0.965} &
   &
   &
   &
   \\ \cline{1-3} \cline{5-6}
\multicolumn{1}{|c|}{all-NVGaze} &
  \multicolumn{1}{c|}{0.960} &
  \multicolumn{1}{c|}{0.930} &
  \multicolumn{1}{c|}{} &
  \multicolumn{1}{c|}{0.974} &
  \multicolumn{1}{c|}{0.965} &
   &
   &
   &
   \\ \cline{1-3} \cline{5-6}
\multicolumn{1}{|c|}{all-UnityEyes} &
  \multicolumn{1}{c|}{0.960} &
  \multicolumn{1}{c|}{0.983} &
  \multicolumn{1}{c|}{} &
  \multicolumn{1}{c|}{0.973} &
  \multicolumn{1}{c|}{0.961} &
   &
   &
   &
   \\ \cline{1-3} \cline{5-6}
\multicolumn{1}{|c|}{all-RITEyes-Gen} &
  \multicolumn{1}{c|}{0.961} &
  \multicolumn{1}{c|}{0.983} &
  \multicolumn{1}{c|}{} &
  \multicolumn{1}{c|}{0.962} &
  \multicolumn{1}{c|}{0.958} &
   &
   &
   &
   \\ \cline{1-3} \cline{5-6}
\multicolumn{1}{|c|}{all-RITEyes-Nat} &
  \multicolumn{1}{c|}{0.960} &
  \multicolumn{1}{c|}{0.984} &
  \multicolumn{1}{c|}{} &
  \multicolumn{1}{c|}{0.968} &
  \multicolumn{1}{c|}{0.928} &
   &
   &
   &
   \\ \cline{1-3} \cline{5-6}
\multicolumn{1}{|c|}{all-LPW} &
  \multicolumn{1}{c|}{0.960} &
  \multicolumn{1}{c|}{0.984} &
  \multicolumn{1}{c|}{} &
  \multicolumn{1}{c|}{0.974} &
  \multicolumn{1}{c|}{0.962} &
   &
   &
   &
   \\ \cline{1-3} \cline{5-6}
\multicolumn{1}{|c|}{all-Santini} &
  \multicolumn{1}{c|}{0.961} &
  \multicolumn{1}{c|}{0.984} &
  \multicolumn{1}{c|}{} &
  \multicolumn{1}{c|}{0.976} &
  \multicolumn{1}{c|}{0.965} &
   &
   &
   &
   \\ \cline{1-3} \cline{5-6}
\multicolumn{1}{|c|}{all-Fuhl} &
  \multicolumn{1}{c|}{0.961} &
  \multicolumn{1}{c|}{0.984} &
  \multicolumn{1}{c|}{} &
  \multicolumn{1}{c|}{0.975} &
  \multicolumn{1}{c|}{0.964} &
   &
   &
   &
   \\ \cline{1-3} \cline{5-6}
\multicolumn{1}{|c|}{all-Swirski} &
  \multicolumn{1}{c|}{0.960} &
  \multicolumn{1}{c|}{0.985} &
  \multicolumn{1}{c|}{} &
  \multicolumn{1}{c|}{0.977} &
  \multicolumn{1}{c|}{0.968} &
   &
   &
   &
   \\ \cline{1-3} \cline{5-6}
\multicolumn{1}{l}{} &
   &
   &
  \multicolumn{1}{l}{} &
  \multicolumn{1}{l}{} &
  \multicolumn{1}{l}{} &
  \multicolumn{1}{l}{} &
  \multicolumn{1}{l}{} &
  \multicolumn{1}{l}{} &
  \multicolumn{1}{l|}{} \\ \cline{1-3} \cline{5-6}
\multicolumn{1}{|c|}{all} &
  \multicolumn{1}{c|}{0.956} &
  \multicolumn{1}{c|}{0.982} &
  \multicolumn{1}{c|}{} &
  \multicolumn{1}{c|}{0.961} &
  \multicolumn{1}{c|}{0.947} &
   &
   &
   &
   \\ \cline{1-3} \cline{5-10} 
\end{tabular}
\end{adjustbox}
\caption{Segmentation results of various generalization tests proposed in Section 3.4. All results represent the IoU metric.}
\label{tab:seg_results_aug_0}
\end{table}

\begin{table}[]
\centering
\begin{adjustbox}{angle=90}
\begin{tabular}{cccccccccc|}
\hline
\multicolumn{1}{|c|}{Train on / Test on} &
  \multicolumn{1}{c|}{OpenEDS} &
  \multicolumn{1}{c|}{NVGaze} &
  \multicolumn{1}{c|}{UnityEyes} &
  \multicolumn{1}{c|}{RITEyes-Gen} &
  \multicolumn{1}{c|}{RITEyes-Nat} &
  \multicolumn{1}{c|}{LPW} &
  \multicolumn{1}{c|}{BAT} &
  \multicolumn{1}{c|}{Fuhl} &
  Swirski \\ \hline
\multicolumn{1}{|c|}{OpenEDS} &
  \multicolumn{1}{c|}{0.683} &
  \multicolumn{1}{c|}{1.925} &
  \multicolumn{1}{c|}{} &
  \multicolumn{1}{c|}{7.970} &
  \multicolumn{1}{c|}{23.625} &
  \multicolumn{1}{c|}{2.462} &
  \multicolumn{1}{c|}{2.406} &
  \multicolumn{1}{c|}{7.736} &
  1.732 \\ \cline{1-3} \cline{5-10} 
\multicolumn{1}{|c|}{NVGaze} &
  \multicolumn{1}{c|}{1.039} &
  \multicolumn{1}{c|}{0.168} &
  \multicolumn{1}{c|}{} &
  \multicolumn{1}{c|}{17.104} &
  \multicolumn{1}{c|}{30.998} &
  \multicolumn{1}{c|}{10.507} &
  \multicolumn{1}{c|}{1.227} &
  \multicolumn{1}{c|}{70.052} &
  4.736 \\ \cline{1-3} \cline{5-10} 
\multicolumn{1}{|c|}{UnityEyes} &
  \multicolumn{1}{c|}{} &
  \multicolumn{1}{c|}{} &
  \multicolumn{1}{c|}{} &
  \multicolumn{1}{c|}{} &
  \multicolumn{1}{c|}{} &
  \multicolumn{1}{c|}{} &
  \multicolumn{1}{c|}{} &
  \multicolumn{1}{c|}{} &
   \\ \cline{1-3} \cline{5-10} 
\multicolumn{1}{|c|}{RITEyes-Gen} &
  \multicolumn{1}{c|}{0.845} &
  \multicolumn{1}{c|}{0.850} &
  \multicolumn{1}{c|}{} &
  \multicolumn{1}{c|}{0.292} &
  \multicolumn{1}{c|}{0.768} &
  \multicolumn{1}{c|}{2.019} &
  \multicolumn{1}{c|}{1.540} &
  \multicolumn{1}{c|}{12.642} &
  0.987 \\ \cline{1-3} \cline{5-10} 
\multicolumn{1}{|c|}{RITEyes-Nat} &
  \multicolumn{1}{c|}{0.818} &
  \multicolumn{1}{c|}{0.767} &
  \multicolumn{1}{c|}{} &
  \multicolumn{1}{c|}{0.305} &
  \multicolumn{1}{c|}{0.407} &
  \multicolumn{1}{c|}{2.122} &
  \multicolumn{1}{c|}{2.366} &
  \multicolumn{1}{c|}{2.985} &
  1.623 \\ \cline{1-3} \cline{5-10} 
\multicolumn{1}{|c|}{LPW} &
  \multicolumn{1}{c|}{1.152} &
  \multicolumn{1}{c|}{3.037} &
  \multicolumn{1}{c|}{} &
  \multicolumn{1}{c|}{12.014} &
  \multicolumn{1}{c|}{32.724} &
  \multicolumn{1}{c|}{1.052} &
  \multicolumn{1}{c|}{1.308} &
  \multicolumn{1}{c|}{17.489} &
  0.788 \\ \cline{1-3} \cline{5-10} 
\multicolumn{1}{|c|}{Santini} &
  \multicolumn{1}{c|}{2.912} &
  \multicolumn{1}{c|}{94.254} &
  \multicolumn{1}{c|}{} &
  \multicolumn{1}{c|}{31.814} &
  \multicolumn{1}{c|}{54.246} &
  \multicolumn{1}{c|}{12.778} &
  \multicolumn{1}{c|}{0.892} &
  \multicolumn{1}{c|}{94.609} &
  2.105 \\ \cline{1-3} \cline{5-10} 
\multicolumn{1}{|c|}{Fuhl} &
  \multicolumn{1}{c|}{1.060} &
  \multicolumn{1}{c|}{0.867} &
  \multicolumn{1}{c|}{} &
  \multicolumn{1}{c|}{4.122} &
  \multicolumn{1}{c|}{8.896} &
  \multicolumn{1}{c|}{3.133} &
  \multicolumn{1}{c|}{1.444} &
  \multicolumn{1}{c|}{1.719} &
  1.731 \\ \cline{1-3} \cline{5-10} 
\multicolumn{1}{|c|}{Swirski} &
  \multicolumn{1}{c|}{4.272} &
  \multicolumn{1}{c|}{169.359} &
  \multicolumn{1}{c|}{} &
  \multicolumn{1}{c|}{35.463} &
  \multicolumn{1}{c|}{71.910} &
  \multicolumn{1}{c|}{13.786} &
  \multicolumn{1}{c|}{2.911} &
  \multicolumn{1}{c|}{98.026} &
  1.844 \\ \cline{1-3} \cline{5-10} 
 &
   &
   &
   &
   &
   &
   &
   &
   &
   \\ \cline{1-3} \cline{5-10} 
\multicolumn{1}{|c|}{all-OpenEDS} &
  \multicolumn{1}{c|}{0.918} &
  \multicolumn{1}{c|}{0.240} &
  \multicolumn{1}{c|}{} &
  \multicolumn{1}{c|}{0.378} &
  \multicolumn{1}{c|}{0.529} &
  \multicolumn{1}{c|}{0.894} &
  \multicolumn{1}{c|}{0.412} &
  \multicolumn{1}{c|}{1.717} &
  0.240 \\ \cline{1-3} \cline{5-10} 
\multicolumn{1}{|c|}{all-NVGaze} &
  \multicolumn{1}{c|}{0.666} &
  \multicolumn{1}{c|}{1.048} &
  \multicolumn{1}{c|}{} &
  \multicolumn{1}{c|}{0.453} &
  \multicolumn{1}{c|}{0.624} &
  \multicolumn{1}{c|}{0.960} &
  \multicolumn{1}{c|}{0.438} &
  \multicolumn{1}{c|}{1.786} &
  0.374 \\ \cline{1-3} \cline{5-10} 
\multicolumn{1}{|c|}{all-UnityEyes} &
  \multicolumn{1}{c|}{0.625} &
  \multicolumn{1}{c|}{0.279} &
  \multicolumn{1}{c|}{} &
  \multicolumn{1}{c|}{0.385} &
  \multicolumn{1}{c|}{0.580} &
  \multicolumn{1}{c|}{0.949} &
  \multicolumn{1}{c|}{0.427} &
  \multicolumn{1}{c|}{1.675} &
  0.274 \\ \cline{1-3} \cline{5-10} 
\multicolumn{1}{|c|}{all-RITEyes-Gen} &
  \multicolumn{1}{c|}{0.659} &
  \multicolumn{1}{c|}{0.326} &
  \multicolumn{1}{c|}{} &
  \multicolumn{1}{c|}{0.692} &
  \multicolumn{1}{c|}{0.778} &
  \multicolumn{1}{c|}{1.062} &
  \multicolumn{1}{c|}{0.622} &
  \multicolumn{1}{c|}{1.789} &
  0.323 \\ \cline{1-3} \cline{5-10} 
\multicolumn{1}{|c|}{all-RITEyes-Nat} &
  \multicolumn{1}{c|}{0.598} &
  \multicolumn{1}{c|}{0.283} &
  \multicolumn{1}{c|}{} &
  \multicolumn{1}{c|}{0.503} &
  \multicolumn{1}{c|}{1.003} &
  \multicolumn{1}{c|}{0.936} &
  \multicolumn{1}{c|}{0.439} &
  \multicolumn{1}{c|}{1.763} &
  0.265 \\ \cline{1-3} \cline{5-10} 
\multicolumn{1}{|c|}{all-LPW} &
  \multicolumn{1}{c|}{0.661} &
  \multicolumn{1}{c|}{0.293} &
  \multicolumn{1}{c|}{} &
  \multicolumn{1}{c|}{0.542} &
  \multicolumn{1}{c|}{0.756} &
  \multicolumn{1}{c|}{1.550} &
  \multicolumn{1}{c|}{0.484} &
  \multicolumn{1}{c|}{1.707} &
  0.414 \\ \cline{1-3} \cline{5-10} 
\multicolumn{1}{|c|}{all-Santini} &
  \multicolumn{1}{c|}{0.606} &
  \multicolumn{1}{c|}{0.265} &
  \multicolumn{1}{c|}{} &
  \multicolumn{1}{c|}{0.343} &
  \multicolumn{1}{c|}{0.515} &
  \multicolumn{1}{c|}{0.896} &
  \multicolumn{1}{c|}{1.543} &
  \multicolumn{1}{c|}{1.689} &
  0.250 \\ \cline{1-3} \cline{5-10} 
\multicolumn{1}{|c|}{all-Fuhl} &
  \multicolumn{1}{c|}{0.597} &
  \multicolumn{1}{c|}{0.272} &
  \multicolumn{1}{c|}{} &
  \multicolumn{1}{c|}{0.414} &
  \multicolumn{1}{c|}{0.579} &
  \multicolumn{1}{c|}{0.920} &
  \multicolumn{1}{c|}{0.500} &
  \multicolumn{1}{c|}{2.627} &
  0.347 \\ \cline{1-3} \cline{5-10} 
\multicolumn{1}{|c|}{all-Swirski} &
  \multicolumn{1}{c|}{0.660} &
  \multicolumn{1}{c|}{0.256} &
  \multicolumn{1}{c|}{} &
  \multicolumn{1}{c|}{0.363} &
  \multicolumn{1}{c|}{0.528} &
  \multicolumn{1}{c|}{0.988} &
  \multicolumn{1}{c|}{0.410} &
  \multicolumn{1}{c|}{1.712} &
  0.980 \\ \cline{1-3} \cline{5-10} 
\multicolumn{1}{l}{} &
  \multicolumn{1}{l}{} &
  \multicolumn{1}{l}{} &
  \multicolumn{1}{l}{} &
  \multicolumn{1}{l}{} &
  \multicolumn{1}{l}{} &
  \multicolumn{1}{l}{} &
  \multicolumn{1}{l}{} &
  \multicolumn{1}{l}{} &
  \multicolumn{1}{l|}{} \\ \cline{1-3} \cline{5-10} 
\multicolumn{1}{|c|}{all} &
  \multicolumn{1}{c|}{0.669} &
  \multicolumn{1}{c|}{0.272} &
  \multicolumn{1}{c|}{} &
  \multicolumn{1}{c|}{0.461} &
  \multicolumn{1}{c|}{0.642} &
  \multicolumn{1}{c|}{1.085} &
  \multicolumn{1}{c|}{0.691} &
  \multicolumn{1}{c|}{1.842} &
  0.463 \\ \cline{1-3} \cline{5-10} 
\end{tabular}
\end{adjustbox}
\caption{Error in pupil center prediction across various generalization tests proposed in Section 3.4. All results are presented in unit pixels.}
\label{tab:pupil_results_aug_0}
\end{table}

\begin{table}[]
\centering
\begin{adjustbox}{angle=90}
\begin{tabular}{cccccccccc|}
\hline
\multicolumn{1}{|c|}{Train on / Test on} &
  \multicolumn{1}{c|}{OpenEDS} &
  \multicolumn{1}{c|}{NVGaze} &
  \multicolumn{1}{c|}{UnityEyes} &
  \multicolumn{1}{c|}{RITEyes-Gen} &
  \multicolumn{1}{c|}{RITEyes-Nat} &
  \multicolumn{1}{c|}{LPW} &
  \multicolumn{1}{c|}{BAT} &
  \multicolumn{1}{c|}{Fuhl} &
  Swirski \\ \hline
\multicolumn{1}{|c|}{OpenEDS} &
  \multicolumn{1}{c|}{1.848} &
  \multicolumn{1}{c|}{5.864} &
  \multicolumn{1}{c|}{11.039} &
  \multicolumn{1}{c|}{14.375} &
  \multicolumn{1}{c|}{31.254} &
   &
   &
   &
   \\ \cline{1-6}
\multicolumn{1}{|c|}{NVGaze} &
  \multicolumn{1}{c|}{4.249} &
  \multicolumn{1}{c|}{0.300} &
  \multicolumn{1}{c|}{9.760} &
  \multicolumn{1}{c|}{15.680} &
  \multicolumn{1}{c|}{31.174} &
   &
   &
   &
   \\ \cline{1-6}
\multicolumn{1}{|c|}{UnityEyes} &
  \multicolumn{1}{c|}{7.438} &
  \multicolumn{1}{c|}{3.458} &
  \multicolumn{1}{c|}{0.651} &
  \multicolumn{1}{c|}{9.790} &
  \multicolumn{1}{c|}{16.928} &
   &
   &
   &
   \\ \cline{1-6}
\multicolumn{1}{|c|}{RITEyes-Gen} &
  \multicolumn{1}{c|}{3.885} &
  \multicolumn{1}{c|}{2.749} &
  \multicolumn{1}{c|}{4.047} &
  \multicolumn{1}{c|}{0.702} &
  \multicolumn{1}{c|}{1.099} &
   &
   &
   &
   \\ \cline{1-6}
\multicolumn{1}{|c|}{RITEyes-Nat} &
  \multicolumn{1}{c|}{4.404} &
  \multicolumn{1}{c|}{2.741} &
  \multicolumn{1}{c|}{2.413} &
  \multicolumn{1}{c|}{0.618} &
  \multicolumn{1}{c|}{0.610} &
   &
   &
   &
   \\ \cline{1-6}
\multicolumn{1}{|c|}{LPW} &
   &
   &
   &
   &
   &
   &
   &
   &
   \\ \cline{1-1}
\multicolumn{1}{|c|}{BAT} &
   &
   &
   &
   &
   &
   &
   &
   &
   \\ \cline{1-1}
\multicolumn{1}{|c|}{Fuhl} &
   &
   &
   &
   &
   &
   &
   &
   &
   \\ \cline{1-1}
\multicolumn{1}{|c|}{Swirski} &
   &
   &
   &
   &
   &
   &
   &
   &
   \\ \cline{1-1}
 &
   &
   &
   &
   &
   &
   &
   &
   &
   \\ \cline{1-6}
\multicolumn{1}{|c|}{all-OpenEDS} &
  \multicolumn{1}{c|}{4.225} &
  \multicolumn{1}{c|}{0.353} &
  \multicolumn{1}{c|}{1.364} &
  \multicolumn{1}{c|}{0.581} &
  \multicolumn{1}{c|}{0.715} &
   &
   &
   &
   \\ \cline{1-6}
\multicolumn{1}{|c|}{all-NVGaze} &
  \multicolumn{1}{c|}{1.860} &
  \multicolumn{1}{c|}{2.731} &
  \multicolumn{1}{c|}{1.643} &
  \multicolumn{1}{c|}{0.721} &
  \multicolumn{1}{c|}{0.853} &
   &
   &
   &
   \\ \cline{1-6}
\multicolumn{1}{|c|}{all-UnityEyes} &
  \multicolumn{1}{c|}{1.903} &
  \multicolumn{1}{c|}{0.367} &
  \multicolumn{1}{c|}{3.582} &
  \multicolumn{1}{c|}{0.663} &
  \multicolumn{1}{c|}{0.800} &
   &
   &
   &
   \\ \cline{1-6}
\multicolumn{1}{|c|}{all-RITEyes-Gen} &
  \multicolumn{1}{c|}{2.060} &
  \multicolumn{1}{c|}{0.462} &
  \multicolumn{1}{c|}{1.697} &
  \multicolumn{1}{c|}{0.997} &
  \multicolumn{1}{c|}{1.064} &
   &
   &
   &
   \\ \cline{1-6}
\multicolumn{1}{|c|}{all-RITEyes-Nat} &
  \multicolumn{1}{c|}{2.121} &
  \multicolumn{1}{c|}{0.430} &
  \multicolumn{1}{c|}{1.928} &
  \multicolumn{1}{c|}{0.893} &
  \multicolumn{1}{c|}{1.513} &
   &
   &
   &
   \\ \cline{1-6}
\multicolumn{1}{|c|}{all-LPW} &
  \multicolumn{1}{c|}{2.097} &
  \multicolumn{1}{c|}{0.413} &
  \multicolumn{1}{c|}{1.501} &
  \multicolumn{1}{c|}{0.716} &
  \multicolumn{1}{c|}{0.874} &
   &
   &
   &
   \\ \cline{1-6}
\multicolumn{1}{|c|}{all-BAT} &
  \multicolumn{1}{c|}{1.858} &
  \multicolumn{1}{c|}{0.399} &
  \multicolumn{1}{c|}{1.430} &
  \multicolumn{1}{c|}{0.698} &
  \multicolumn{1}{c|}{0.829} &
   &
   &
   &
   \\ \cline{1-6}
\multicolumn{1}{|c|}{all-Fuhl} &
  \multicolumn{1}{c|}{2.037} &
  \multicolumn{1}{c|}{0.405} &
  \multicolumn{1}{c|}{1.620} &
  \multicolumn{1}{c|}{0.645} &
  \multicolumn{1}{c|}{0.783} &
   &
   &
   &
   \\ \cline{1-6}
\multicolumn{1}{|c|}{all-Swirski} &
  \multicolumn{1}{c|}{1.856} &
  \multicolumn{1}{c|}{0.441} &
  \multicolumn{1}{c|}{1.427} &
  \multicolumn{1}{c|}{0.689} &
  \multicolumn{1}{c|}{0.822} &
   &
   &
   &
   \\ \cline{1-6}
\multicolumn{1}{l}{} &
  \multicolumn{1}{l}{} &
  \multicolumn{1}{l}{} &
  \multicolumn{1}{l}{} &
  \multicolumn{1}{l}{} &
  \multicolumn{1}{l}{} &
  \multicolumn{1}{l}{} &
  \multicolumn{1}{l}{} &
  \multicolumn{1}{l}{} &
  \multicolumn{1}{l|}{} \\ \cline{1-6}
\multicolumn{1}{|c|}{all} &
  \multicolumn{1}{c|}{2.171} &
  \multicolumn{1}{c|}{0.441} &
  \multicolumn{1}{c|}{1.460} &
  \multicolumn{1}{c|}{0.883} &
  \multicolumn{1}{c|}{1.033} &
   &
   &
   &
   \\ \hline
\end{tabular}
\end{adjustbox}
\caption{Error in iris center prediction across various generalization tests proposed in Section 3.4. All results are presented in unit pixels.}
\label{tab:iris_results_aug_0}
\end{table}
\begin{table}[]
\centering
\begin{adjustbox}{angle=90}
\begin{tabular}{cccccccccc|}
\hline
\multicolumn{1}{|c|}{Train on / Test on} &
  \multicolumn{1}{c|}{OpenEDS} &
  \multicolumn{1}{c|}{NVGaze} &
  \multicolumn{1}{c|}{UnityEyes} &
  \multicolumn{1}{c|}{RITEyes-Gen} &
  \multicolumn{1}{c|}{RITEyes-Nat} &
  \multicolumn{1}{c|}{LPW} &
  \multicolumn{1}{c|}{BAT} &
  \multicolumn{1}{c|}{Fuhl} &
  Swirski \\ \hline
\multicolumn{1}{|c|}{OpenEDS} &
  \multicolumn{1}{c|}{0.956} &
  \multicolumn{1}{c|}{0.572} &
  \multicolumn{1}{c|}{} &
  \multicolumn{1}{c|}{0.550} &
  \multicolumn{1}{c|}{0.427} &
   &
   &
   &
   \\ \cline{1-3} \cline{5-6}
\multicolumn{1}{|c|}{NVGaze} &
  \multicolumn{1}{c|}{0.664} &
  \multicolumn{1}{c|}{0.986} &
  \multicolumn{1}{c|}{} &
  \multicolumn{1}{c|}{0.570} &
  \multicolumn{1}{c|}{0.355} &
   &
   &
   &
   \\ \cline{1-3} \cline{5-6}
\multicolumn{1}{|c|}{UnityEyes} &
  \multicolumn{1}{c|}{} &
  \multicolumn{1}{c|}{} &
  \multicolumn{1}{c|}{} &
  \multicolumn{1}{c|}{} &
  \multicolumn{1}{c|}{} &
   &
   &
   &
   \\ \cline{1-3} \cline{5-6}
\multicolumn{1}{|c|}{RITEyes-Gen} &
  \multicolumn{1}{c|}{0.713} &
  \multicolumn{1}{c|}{0.883} &
  \multicolumn{1}{c|}{} &
  \multicolumn{1}{c|}{0.960} &
  \multicolumn{1}{c|}{0.932} &
   &
   &
   &
   \\ \cline{1-3} \cline{5-6}
\multicolumn{1}{|c|}{RITEyes-Nat} &
  \multicolumn{1}{c|}{0.812} &
  \multicolumn{1}{c|}{0.917} &
  \multicolumn{1}{c|}{} &
  \multicolumn{1}{c|}{0.954} &
  \multicolumn{1}{c|}{0.948} &
   &
   &
   &
   \\ \cline{1-3} \cline{5-6}
\multicolumn{1}{|c|}{LPW} &
   &
   &
   &
   &
  \multicolumn{1}{c|}{} &
   &
   &
   &
   \\ \cline{1-1}
\multicolumn{1}{|c|}{Santini} &
   &
   &
   &
   &
  \multicolumn{1}{c|}{} &
   &
   &
   &
   \\ \cline{1-1}
\multicolumn{1}{|c|}{Fuhl} &
   &
   &
   &
   &
  \multicolumn{1}{c|}{} &
   &
   &
   &
   \\ \cline{1-1}
\multicolumn{1}{|c|}{Swirski} &
   &
   &
   &
   &
  \multicolumn{1}{c|}{} &
   &
   &
   &
   \\ \cline{1-6}
 &
   &
   &
   &
   &
   &
   &
   &
   &
   \\ \cline{1-3} \cline{5-6}
\multicolumn{1}{|c|}{all-OpenEDS} &
  \multicolumn{1}{c|}{0.688} &
  \multicolumn{1}{c|}{0.486} &
  \multicolumn{1}{c|}{} &
  \multicolumn{1}{c|}{0.822} &
  \multicolumn{1}{c|}{0.850} &
   &
   &
   &
   \\ \cline{1-3} \cline{5-6}
\multicolumn{1}{|c|}{all-NVGaze} &
  \multicolumn{1}{c|}{0.920} &
  \multicolumn{1}{c|}{0.428} &
  \multicolumn{1}{c|}{} &
  \multicolumn{1}{c|}{0.729} &
  \multicolumn{1}{c|}{0.757} &
   &
   &
   &
   \\ \cline{1-3} \cline{5-6}
\multicolumn{1}{|c|}{all-UnityEyes} &
  \multicolumn{1}{c|}{0.949} &
  \multicolumn{1}{c|}{0.950} &
  \multicolumn{1}{c|}{} &
  \multicolumn{1}{c|}{0.950} &
  \multicolumn{1}{c|}{0.937} &
   &
   &
   &
   \\ \cline{1-3} \cline{5-6}
\multicolumn{1}{|c|}{all-RITEyes-Gen} &
  \multicolumn{1}{c|}{0.933} &
  \multicolumn{1}{c|}{0.442} &
  \multicolumn{1}{c|}{} &
  \multicolumn{1}{c|}{0.685} &
  \multicolumn{1}{c|}{0.787} &
   &
   &
   &
   \\ \cline{1-3} \cline{5-6}
\multicolumn{1}{|c|}{all-RITEyes-Nat} &
  \multicolumn{1}{c|}{0.939} &
  \multicolumn{1}{c|}{0.466} &
  \multicolumn{1}{c|}{} &
  \multicolumn{1}{c|}{0.802} &
  \multicolumn{1}{c|}{0.698} &
   &
   &
   &
   \\ \cline{1-3} \cline{5-6}
\multicolumn{1}{|c|}{all-LPW} &
  \multicolumn{1}{c|}{0.952} &
  \multicolumn{1}{c|}{0.967} &
  \multicolumn{1}{c|}{} &
  \multicolumn{1}{c|}{0.954} &
  \multicolumn{1}{c|}{0.942} &
   &
   &
   &
   \\ \cline{1-3} \cline{5-6}
\multicolumn{1}{|c|}{all-Santini} &
  \multicolumn{1}{c|}{0.952} &
  \multicolumn{1}{c|}{0.963} &
  \multicolumn{1}{c|}{} &
  \multicolumn{1}{c|}{0.957} &
  \multicolumn{1}{c|}{0.946} &
   &
   &
   &
   \\ \cline{1-3} \cline{5-6}
\multicolumn{1}{|c|}{all-Fuhl} &
  \multicolumn{1}{c|}{0.949} &
  \multicolumn{1}{c|}{0.958} &
  \multicolumn{1}{c|}{} &
  \multicolumn{1}{c|}{0.955} &
  \multicolumn{1}{c|}{0.943} &
   &
   &
   &
   \\ \cline{1-3} \cline{5-6}
\multicolumn{1}{|c|}{all-Swirski} &
  \multicolumn{1}{c|}{0.950} &
  \multicolumn{1}{c|}{0.954} &
  \multicolumn{1}{c|}{} &
  \multicolumn{1}{c|}{0.952} &
  \multicolumn{1}{c|}{0.940} &
   &
   &
   &
   \\ \cline{1-3} \cline{5-6}
\multicolumn{1}{l}{} &
   &
   &
  \multicolumn{1}{l}{} &
  \multicolumn{1}{l}{} &
  \multicolumn{1}{l}{} &
  \multicolumn{1}{l}{} &
  \multicolumn{1}{l}{} &
  \multicolumn{1}{l}{} &
  \multicolumn{1}{l|}{} \\ \cline{1-3} \cline{5-6}
\multicolumn{1}{|c|}{all} &
  \multicolumn{1}{c|}{0.956} &
  \multicolumn{1}{c|}{0.982} &
  \multicolumn{1}{c|}{} &
  \multicolumn{1}{c|}{0.961} &
  \multicolumn{1}{c|}{0.947} &
   &
   &
   &
   \\ \cline{1-3} \cline{5-10} 
\end{tabular}
\end{adjustbox}
\caption{Segmentation results of various generalization tests proposed in Section 3.4 when utilizing Batch Normalization. All results represent the IoU metric.}
\label{tab:BN_seg_results_aug_0}
\end{table}

\begin{table}[]
\centering
\begin{adjustbox}{angle=90}
\begin{tabular}{cccccccccc|}
\hline
\multicolumn{1}{|c|}{Train on / Test on} &
  \multicolumn{1}{c|}{OpenEDS} &
  \multicolumn{1}{c|}{NVGaze} &
  \multicolumn{1}{c|}{UnityEyes} &
  \multicolumn{1}{c|}{RITEyes-Gen} &
  \multicolumn{1}{c|}{RITEyes-Nat} &
  \multicolumn{1}{c|}{LPW} &
  \multicolumn{1}{c|}{BAT} &
  \multicolumn{1}{c|}{Fuhl} &
  Swirski \\ \hline
\multicolumn{1}{|c|}{OpenEDS} &
  \multicolumn{1}{c|}{0.614} &
  \multicolumn{1}{c|}{2.102} &
  \multicolumn{1}{c|}{} &
  \multicolumn{1}{c|}{8.346} &
  \multicolumn{1}{c|}{18.620} &
  \multicolumn{1}{c|}{1.349} &
  \multicolumn{1}{c|}{1.951} &
  \multicolumn{1}{c|}{4.920} &
  1.583 \\ \cline{1-3} \cline{5-10} 
\multicolumn{1}{|c|}{NVGaze} &
  \multicolumn{1}{c|}{1.031} &
  \multicolumn{1}{c|}{0.201} &
  \multicolumn{1}{c|}{} &
  \multicolumn{1}{c|}{13.952} &
  \multicolumn{1}{c|}{61.093} &
  \multicolumn{1}{c|}{4.483} &
  \multicolumn{1}{c|}{2.139} &
  \multicolumn{1}{c|}{67.145} &
  4.146 \\ \cline{1-3} \cline{5-10} 
\multicolumn{1}{|c|}{UnityEyes} &
  \multicolumn{1}{c|}{} &
  \multicolumn{1}{c|}{} &
  \multicolumn{1}{c|}{} &
  \multicolumn{1}{c|}{} &
  \multicolumn{1}{c|}{} &
  \multicolumn{1}{c|}{} &
  \multicolumn{1}{c|}{} &
  \multicolumn{1}{c|}{} &
   \\ \cline{1-3} \cline{5-10} 
\multicolumn{1}{|c|}{RITEyes-Gen} &
  \multicolumn{1}{c|}{0.732} &
  \multicolumn{1}{c|}{1.262} &
  \multicolumn{1}{c|}{} &
  \multicolumn{1}{c|}{0.302} &
  \multicolumn{1}{c|}{0.720} &
  \multicolumn{1}{c|}{1.735} &
  \multicolumn{1}{c|}{2.375} &
  \multicolumn{1}{c|}{32.363} &
  1.160 \\ \cline{1-3} \cline{5-10} 
\multicolumn{1}{|c|}{RITEyes-Nat} &
  \multicolumn{1}{c|}{0.801} &
  \multicolumn{1}{c|}{0.681} &
  \multicolumn{1}{c|}{} &
  \multicolumn{1}{c|}{0.355} &
  \multicolumn{1}{c|}{0.481} &
  \multicolumn{1}{c|}{1.581} &
  \multicolumn{1}{c|}{2.243} &
  \multicolumn{1}{c|}{6.457} &
  2.001 \\ \cline{1-3} \cline{5-10} 
\multicolumn{1}{|c|}{LPW} &
  \multicolumn{1}{c|}{0.737} &
  \multicolumn{1}{c|}{1.448} &
  \multicolumn{1}{c|}{} &
  \multicolumn{1}{c|}{3.205} &
  \multicolumn{1}{c|}{8.256} &
  \multicolumn{1}{c|}{0.638} &
  \multicolumn{1}{c|}{1.694} &
  \multicolumn{1}{c|}{5.471} &
  1.022 \\ \cline{1-3} \cline{5-10} 
\multicolumn{1}{|c|}{Santini} &
  \multicolumn{1}{c|}{1.740} &
  \multicolumn{1}{c|}{1.838} &
  \multicolumn{1}{c|}{} &
  \multicolumn{1}{c|}{3.256} &
  \multicolumn{1}{c|}{4.458} &
  \multicolumn{1}{c|}{2.634} &
  \multicolumn{1}{c|}{0.494} &
  \multicolumn{1}{c|}{3.994} &
  1.633 \\ \cline{1-3} \cline{5-10} 
\multicolumn{1}{|c|}{Fuhl} &
  \multicolumn{1}{c|}{1.313} &
  \multicolumn{1}{c|}{1.796} &
  \multicolumn{1}{c|}{} &
  \multicolumn{1}{c|}{3.469} &
  \multicolumn{1}{c|}{5.272} &
  \multicolumn{1}{c|}{2.696} &
  \multicolumn{1}{c|}{1.045} &
  \multicolumn{1}{c|}{1.724} &
  1.740 \\ \cline{1-3} \cline{5-10} 
\multicolumn{1}{|c|}{Swirski} &
  \multicolumn{1}{c|}{2.003} &
  \multicolumn{1}{c|}{2.723} &
  \multicolumn{1}{c|}{} &
  \multicolumn{1}{c|}{3.818} &
  \multicolumn{1}{c|}{5.200} &
  \multicolumn{1}{c|}{4.189} &
  \multicolumn{1}{c|}{0.747} &
  \multicolumn{1}{c|}{7.593} &
  0.616 \\ \cline{1-3} \cline{5-10} 
 &
   &
   &
   &
   &
   &
   &
   &
   &
   \\ \cline{1-3} \cline{5-10} 
\multicolumn{1}{|c|}{all-OpenEDS} &
  \multicolumn{1}{c|}{1.387} &
  \multicolumn{1}{c|}{0.637} &
  \multicolumn{1}{c|}{} &
  \multicolumn{1}{c|}{0.717} &
  \multicolumn{1}{c|}{0.924} &
  \multicolumn{1}{c|}{1.260} &
  \multicolumn{1}{c|}{0.688} &
  \multicolumn{1}{c|}{1.771} &
  0.437 \\ \cline{1-3} \cline{5-10} 
\multicolumn{1}{|c|}{all-NVGaze} &
  \multicolumn{1}{c|}{1.088} &
  \multicolumn{1}{c|}{1.196} &
  \multicolumn{1}{c|}{} &
  \multicolumn{1}{c|}{0.767} &
  \multicolumn{1}{c|}{0.897} &
  \multicolumn{1}{c|}{1.089} &
  \multicolumn{1}{c|}{0.585} &
  \multicolumn{1}{c|}{2.060} &
  0.704 \\ \cline{1-3} \cline{5-10} 
\multicolumn{1}{|c|}{all-UnityEyes} &
  \multicolumn{1}{c|}{1.091} &
  \multicolumn{1}{c|}{0.660} &
  \multicolumn{1}{c|}{} &
  \multicolumn{1}{c|}{0.544} &
  \multicolumn{1}{c|}{0.674} &
  \multicolumn{1}{c|}{0.826} &
  \multicolumn{1}{c|}{0.529} &
  \multicolumn{1}{c|}{1.798} &
  0.392 \\ \cline{1-3} \cline{5-10} 
\multicolumn{1}{|c|}{all-RITEyes-Gen} &
  \multicolumn{1}{c|}{1.614} &
  \multicolumn{1}{c|}{0.820} &
  \multicolumn{1}{c|}{} &
  \multicolumn{1}{c|}{0.957} &
  \multicolumn{1}{c|}{1.051} &
  \multicolumn{1}{c|}{1.166} &
  \multicolumn{1}{c|}{0.492} &
  \multicolumn{1}{c|}{1.701} &
  0.447 \\ \cline{1-3} \cline{5-10} 
\multicolumn{1}{|c|}{all-RITEyes-Nat} &
  \multicolumn{1}{c|}{1.242} &
  \multicolumn{1}{c|}{0.725} &
  \multicolumn{1}{c|}{} &
  \multicolumn{1}{c|}{0.786} &
  \multicolumn{1}{c|}{1.213} &
  \multicolumn{1}{c|}{0.986} &
  \multicolumn{1}{c|}{0.623} &
  \multicolumn{1}{c|}{1.918} &
  0.530 \\ \cline{1-3} \cline{5-10} 
\multicolumn{1}{|c|}{all-LPW} &
  \multicolumn{1}{c|}{1.071} &
  \multicolumn{1}{c|}{0.353} &
  \multicolumn{1}{c|}{} &
  \multicolumn{1}{c|}{0.462} &
  \multicolumn{1}{c|}{0.600} &
  \multicolumn{1}{c|}{1.310} &
  \multicolumn{1}{c|}{0.419} &
  \multicolumn{1}{c|}{1.718} &
  0.265 \\ \cline{1-3} \cline{5-10} 
\multicolumn{1}{|c|}{all-Santini} &
  \multicolumn{1}{c|}{0.829} &
  \multicolumn{1}{c|}{0.306} &
  \multicolumn{1}{c|}{} &
  \multicolumn{1}{c|}{0.499} &
  \multicolumn{1}{c|}{0.657} &
  \multicolumn{1}{c|}{0.822} &
  \multicolumn{1}{c|}{1.390} &
  \multicolumn{1}{c|}{1.937} &
  0.328 \\ \cline{1-3} \cline{5-10} 
\multicolumn{1}{|c|}{all-Fuhl} &
  \multicolumn{1}{c|}{0.823} &
  \multicolumn{1}{c|}{0.369} &
  \multicolumn{1}{c|}{} &
  \multicolumn{1}{c|}{0.487} &
  \multicolumn{1}{c|}{0.643} &
  \multicolumn{1}{c|}{0.791} &
  \multicolumn{1}{c|}{0.748} &
  \multicolumn{1}{c|}{2.374} &
  0.448 \\ \cline{1-3} \cline{5-10} 
\multicolumn{1}{|c|}{all-Swirski} &
  \multicolumn{1}{c|}{1.113} &
  \multicolumn{1}{c|}{0.491} &
  \multicolumn{1}{c|}{} &
  \multicolumn{1}{c|}{0.435} &
  \multicolumn{1}{c|}{0.577} &
  \multicolumn{1}{c|}{0.855} &
  \multicolumn{1}{c|}{0.478} &
  \multicolumn{1}{c|}{1.777} &
  0.533 \\ \cline{1-3} \cline{5-10} 
\multicolumn{1}{l}{} &
  \multicolumn{1}{l}{} &
  \multicolumn{1}{l}{} &
  \multicolumn{1}{l}{} &
  \multicolumn{1}{l}{} &
  \multicolumn{1}{l}{} &
  \multicolumn{1}{l}{} &
  \multicolumn{1}{l}{} &
  \multicolumn{1}{l}{} &
  \multicolumn{1}{l|}{} \\ \cline{1-3} \cline{5-10} 
\multicolumn{1}{|c|}{all} &
  \multicolumn{1}{c|}{1.190} &
  \multicolumn{1}{c|}{0.645} &
  \multicolumn{1}{c|}{} &
  \multicolumn{1}{c|}{0.673} &
  \multicolumn{1}{c|}{0.835} &
  \multicolumn{1}{c|}{1.156} &
  \multicolumn{1}{c|}{0.934} &
  \multicolumn{1}{c|}{1.915} &
  0.732 \\ \cline{1-3} \cline{5-10} 
\end{tabular}
\end{adjustbox}
\caption{Error in pupil center prediction across various generalization tests proposed in Section 3.4 when utilizing Batch Normalization. All results are presented in unit pixels.}
\label{tab:BN_pupil_results_aug_0}
\end{table}

\begin{table}[]
\centering
\begin{adjustbox}{angle=90}
\begin{tabular}{cccccccccc|}
\hline
\multicolumn{1}{|c|}{Train on / Test on} &
  \multicolumn{1}{c|}{OpenEDS} &
  \multicolumn{1}{c|}{NVGaze} &
  \multicolumn{1}{c|}{UnityEyes} &
  \multicolumn{1}{c|}{RITEyes-Gen} &
  \multicolumn{1}{c|}{RITEyes-Nat} &
  \multicolumn{1}{c|}{LPW} &
  \multicolumn{1}{c|}{BAT} &
  \multicolumn{1}{c|}{Fuhl} &
  Swirski \\ \hline
\multicolumn{1}{|c|}{OpenEDS} &
  \multicolumn{1}{c|}{1.782} &
  \multicolumn{1}{c|}{5.021} &
  \multicolumn{1}{c|}{27.025} &
  \multicolumn{1}{c|}{21.102} &
  \multicolumn{1}{c|}{31.861} &
   &
   &
   &
   \\ \cline{1-6}
\multicolumn{1}{|c|}{NVGaze} &
  \multicolumn{1}{c|}{5.354} &
  \multicolumn{1}{c|}{0.268} &
  \multicolumn{1}{c|}{14.095} &
  \multicolumn{1}{c|}{16.955} &
  \multicolumn{1}{c|}{44.660} &
   &
   &
   &
   \\ \cline{1-6}
\multicolumn{1}{|c|}{UnityEyes} &
  \multicolumn{1}{c|}{4.629} &
  \multicolumn{1}{c|}{2.340} &
  \multicolumn{1}{c|}{0.807} &
  \multicolumn{1}{c|}{2.251} &
  \multicolumn{1}{c|}{2.988} &
   &
   &
   &
   \\ \cline{1-6}
\multicolumn{1}{|c|}{RITEyes-Gen} &
  \multicolumn{1}{c|}{4.568} &
  \multicolumn{1}{c|}{2.687} &
  \multicolumn{1}{c|}{10.048} &
  \multicolumn{1}{c|}{0.598} &
  \multicolumn{1}{c|}{0.928} &
   &
   &
   &
   \\ \cline{1-6}
\multicolumn{1}{|c|}{RITEyes-Nat} &
  \multicolumn{1}{c|}{4.861} &
  \multicolumn{1}{c|}{2.564} &
  \multicolumn{1}{c|}{5.654} &
  \multicolumn{1}{c|}{0.743} &
  \multicolumn{1}{c|}{0.756} &
   &
   &
   &
   \\ \cline{1-6}
\multicolumn{1}{|c|}{LPW} &
   &
   &
   &
   &
   &
   &
   &
   &
   \\ \cline{1-1}
\multicolumn{1}{|c|}{BAT} &
   &
   &
   &
   &
   &
   &
   &
   &
   \\ \cline{1-1}
\multicolumn{1}{|c|}{Fuhl} &
   &
   &
   &
   &
   &
   &
   &
   &
   \\ \cline{1-1}
\multicolumn{1}{|c|}{Swirski} &
   &
   &
   &
   &
   &
   &
   &
   &
   \\ \cline{1-1}
 &
   &
   &
   &
   &
   &
   &
   &
   &
   \\ \cline{1-6}
\multicolumn{1}{|c|}{all-OpenEDS} &
  \multicolumn{1}{c|}{3.823} &
  \multicolumn{1}{c|}{1.610} &
  \multicolumn{1}{c|}{1.920} &
  \multicolumn{1}{c|}{1.111} &
  \multicolumn{1}{c|}{1.268} &
   &
   &
   &
   \\ \cline{1-6}
\multicolumn{1}{|c|}{all-NVGaze} &
  \multicolumn{1}{c|}{2.828} &
  \multicolumn{1}{c|}{2.999} &
  \multicolumn{1}{c|}{2.161} &
  \multicolumn{1}{c|}{1.036} &
  \multicolumn{1}{c|}{1.124} &
   &
   &
   &
   \\ \cline{1-6}
\multicolumn{1}{|c|}{all-UnityEyes} &
  \multicolumn{1}{c|}{2.381} &
  \multicolumn{1}{c|}{1.526} &
  \multicolumn{1}{c|}{3.848} &
  \multicolumn{1}{c|}{1.181} &
  \multicolumn{1}{c|}{1.293} &
   &
   &
   &
   \\ \cline{1-6}
\multicolumn{1}{|c|}{all-RITEyes-Gen} &
  \multicolumn{1}{c|}{2.793} &
  \multicolumn{1}{c|}{1.789} &
  \multicolumn{1}{c|}{1.782} &
  \multicolumn{1}{c|}{1.074} &
  \multicolumn{1}{c|}{1.140} &
   &
   &
   &
   \\ \cline{1-6}
\multicolumn{1}{|c|}{all-RITEyes-Nat} &
  \multicolumn{1}{c|}{2.824} &
  \multicolumn{1}{c|}{2.255} &
  \multicolumn{1}{c|}{1.687} &
  \multicolumn{1}{c|}{1.086} &
  \multicolumn{1}{c|}{1.476} &
   &
   &
   &
   \\ \cline{1-6}
\multicolumn{1}{|c|}{all-LPW} &
  \multicolumn{1}{c|}{2.347} &
  \multicolumn{1}{c|}{0.763} &
  \multicolumn{1}{c|}{1.774} &
  \multicolumn{1}{c|}{0.879} &
  \multicolumn{1}{c|}{0.982} &
   &
   &
   &
   \\ \cline{1-6}
\multicolumn{1}{|c|}{all-BAT} &
  \multicolumn{1}{c|}{2.603} &
  \multicolumn{1}{c|}{0.987} &
  \multicolumn{1}{c|}{2.286} &
  \multicolumn{1}{c|}{1.026} &
  \multicolumn{1}{c|}{1.158} &
   &
   &
   &
   \\ \cline{1-6}
\multicolumn{1}{|c|}{all-Fuhl} &
  \multicolumn{1}{c|}{2.721} &
  \multicolumn{1}{c|}{0.847} &
  \multicolumn{1}{c|}{2.047} &
  \multicolumn{1}{c|}{0.858} &
  \multicolumn{1}{c|}{0.957} &
   &
   &
   &
   \\ \cline{1-6}
\multicolumn{1}{|c|}{all-Swirski} &
  \multicolumn{1}{c|}{2.301} &
  \multicolumn{1}{c|}{1.092} &
  \multicolumn{1}{c|}{2.241} &
  \multicolumn{1}{c|}{1.034} &
  \multicolumn{1}{c|}{1.115} &
   &
   &
   &
   \\ \cline{1-6}
\multicolumn{1}{l}{} &
  \multicolumn{1}{l}{} &
  \multicolumn{1}{l}{} &
  \multicolumn{1}{l}{} &
  \multicolumn{1}{l}{} &
  \multicolumn{1}{l}{} &
  \multicolumn{1}{l}{} &
  \multicolumn{1}{l}{} &
  \multicolumn{1}{l}{} &
  \multicolumn{1}{l|}{} \\ \cline{1-6}
\multicolumn{1}{|c|}{all} &
  \multicolumn{1}{c|}{2.171} &
  \multicolumn{1}{c|}{0.441} &
  \multicolumn{1}{c|}{1.460} &
  \multicolumn{1}{c|}{0.883} &
  \multicolumn{1}{c|}{1.033} &
   &
   &
   &
   \\ \hline
\end{tabular}
\end{adjustbox}
\caption{Error in iris center prediction across various generalization tests proposed in Section 3.4 when utilizing Batch Normalization. All results are presented in unit pixels.}
\label{tab:BN_iris_results_aug_0}
\end{table}

\begin{figure}
    \centering
    \includegraphics[width=\linewidth]{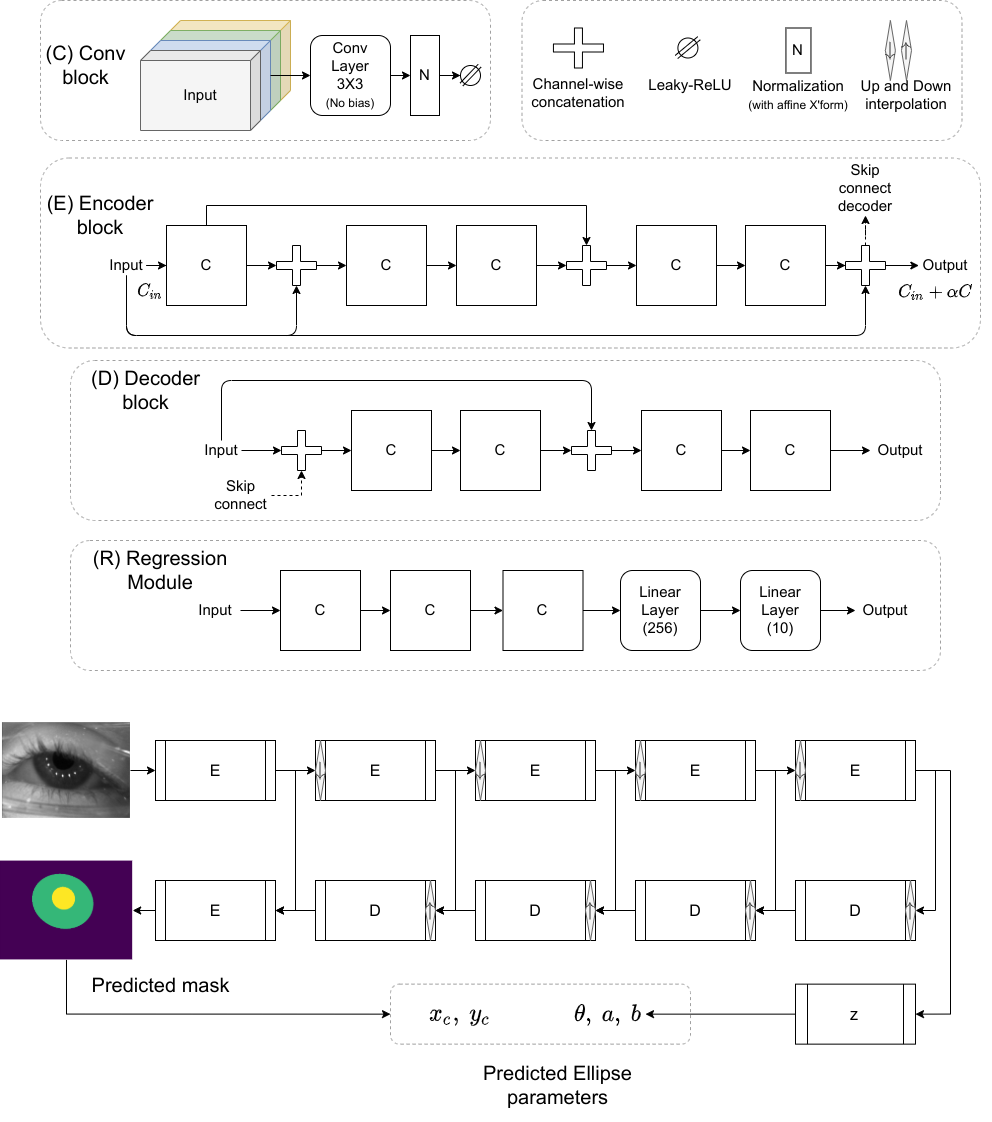}
    \caption{DenseElNet architecture adapted from EllSeg. The number of parameters in the encoder are controlled by a base channel size of $C$ and a growth rate of $\alpha$.}
    \label{fig:EllSeg_gen_arch}
\end{figure}

\begin{figure}
    \centering
    \includegraphics[height=\columnwidth]{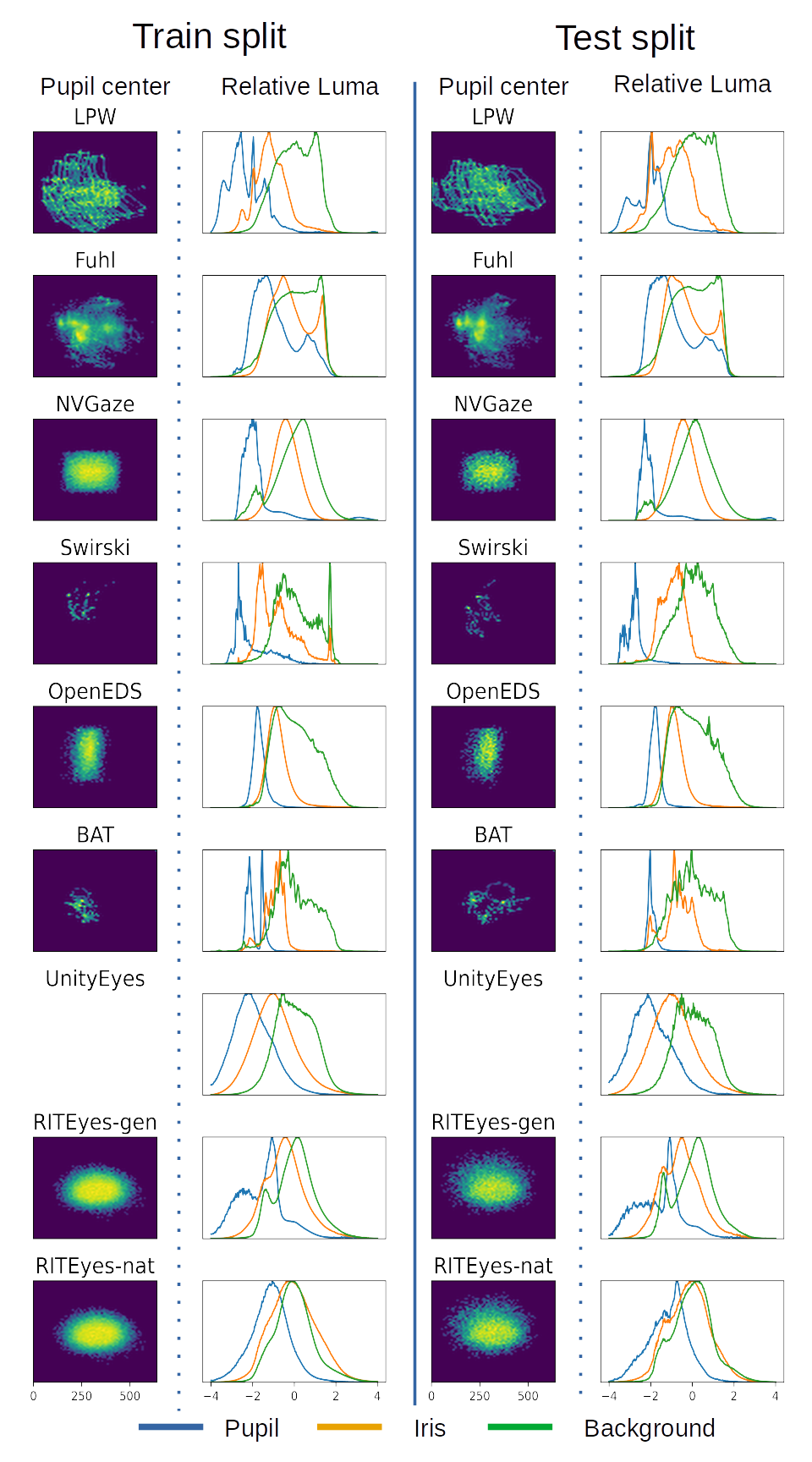}
    \caption{Pupil center (in pixels) and normalized luminance distribution (in Z-scores) of each eye part across all datasets utilized in our experiments (see Table 1). The left and right columns contain statistics from the training and test images for each domain respectively. Due to partial annotations present in some datasets, we leverage the all-vs-one model predictions to segment eye images from all datasets into pupil, iris and background segments. Luminance statistics are then accumulated from the predicted segmentation map.}
    \label{fig:overall_stats}
\end{figure}

\begin{figure}
\centering
\begin{subfigure}{.48\textwidth}
  \centering
  \captionsetup{singlelinecheck=off, margin={0.5cm, 0.5cm}, format=hang}
  \includegraphics[width=\linewidth]{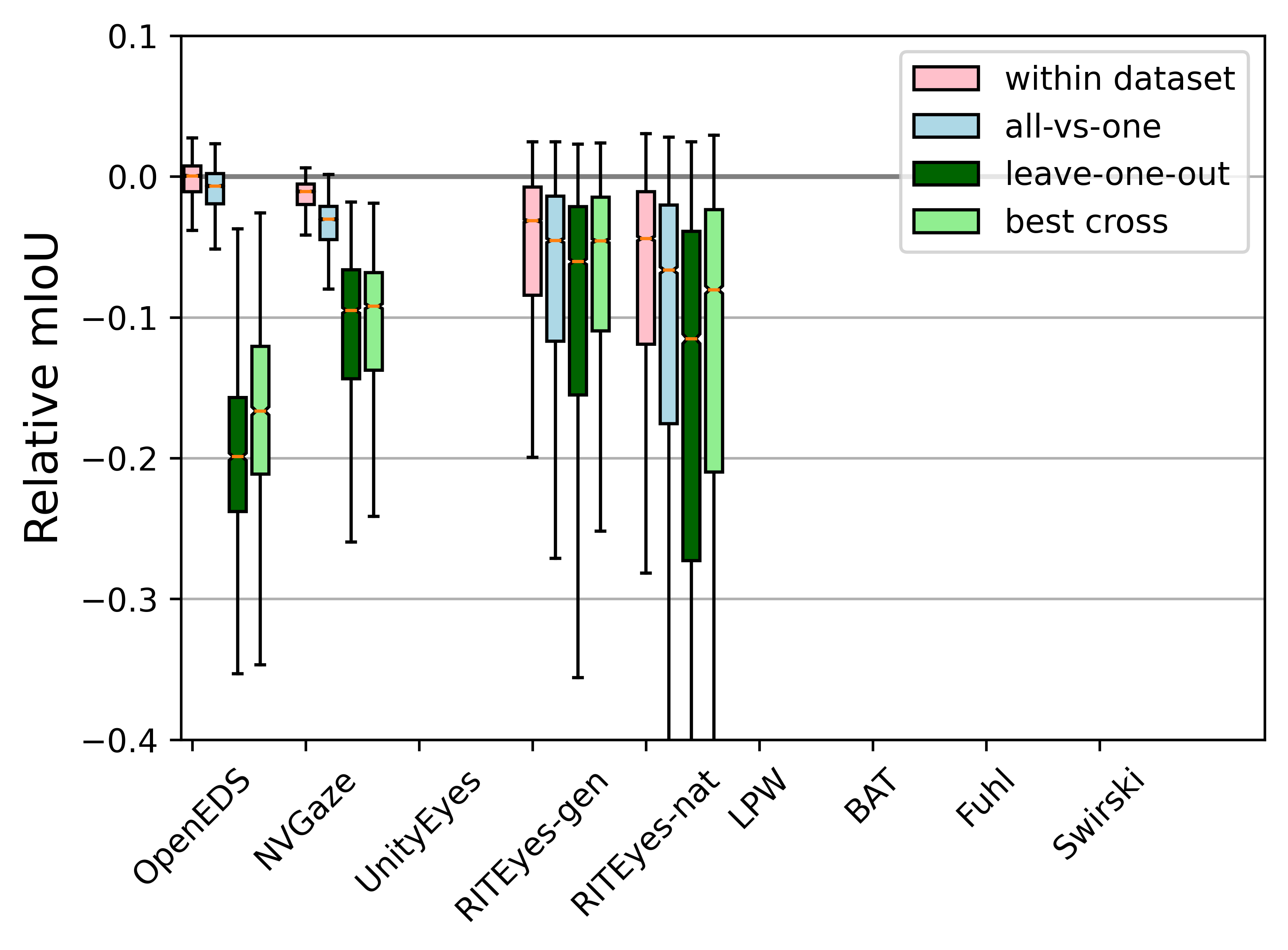}  
  \caption{Performance following reduction in network complexity measured as the mIoU metric relative to training in the absence of augmentation. Positive values indicate a performance improvement.}
\end{subfigure}
\begin{subfigure}{.48\textwidth}
  \centering
  \captionsetup{singlelinecheck=off, margin={0.5cm, 0.5cm}, format=hang}
  \includegraphics[width=\linewidth]{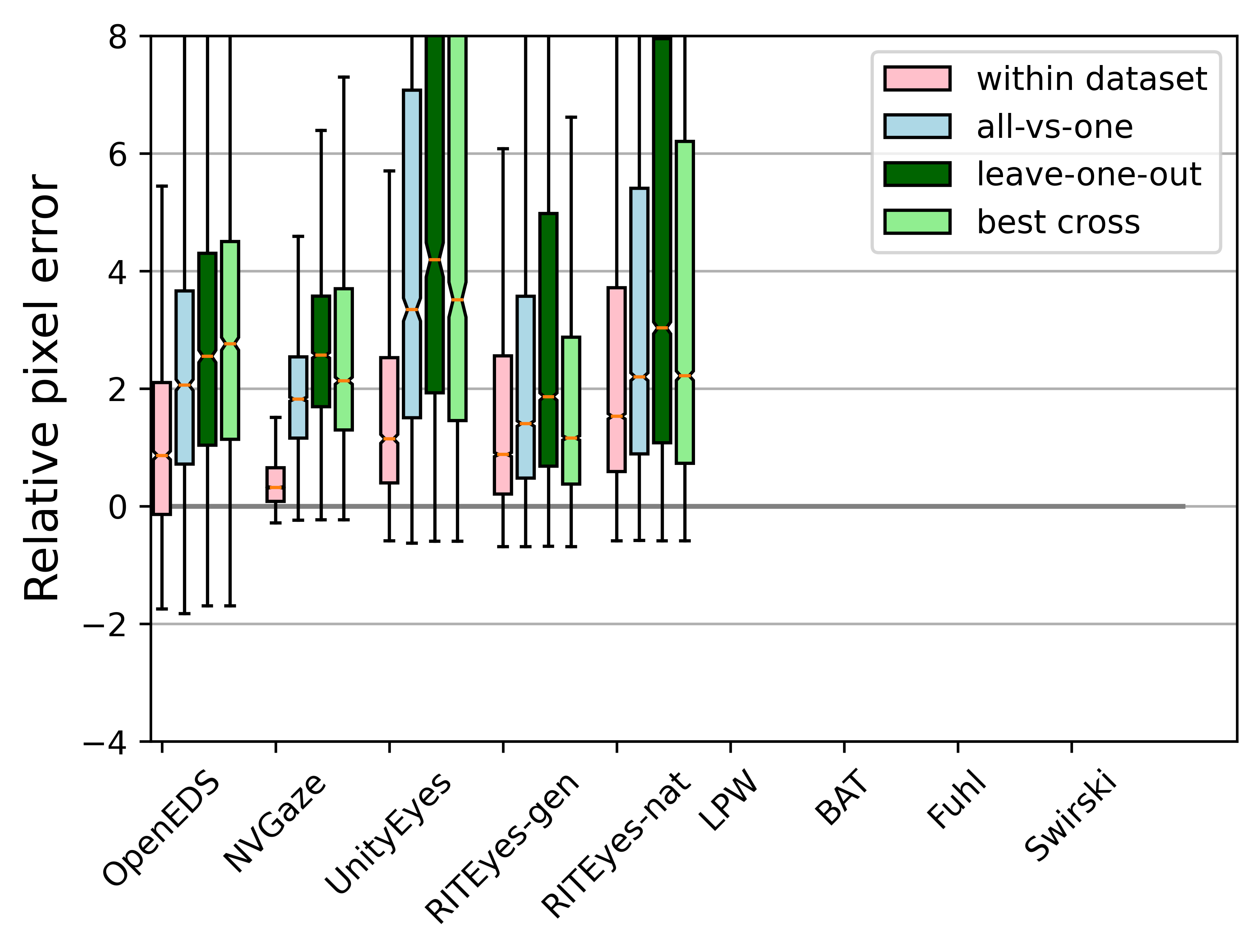}
  \caption{Performance following reduction in network complexity measured as error in iris center, $e_i$, in pixels. Negative values indicate a performance improvement.}
\end{subfigure}

\begin{subfigure}{.8\textwidth}
  \centering
  \includegraphics[width=0.9\linewidth]{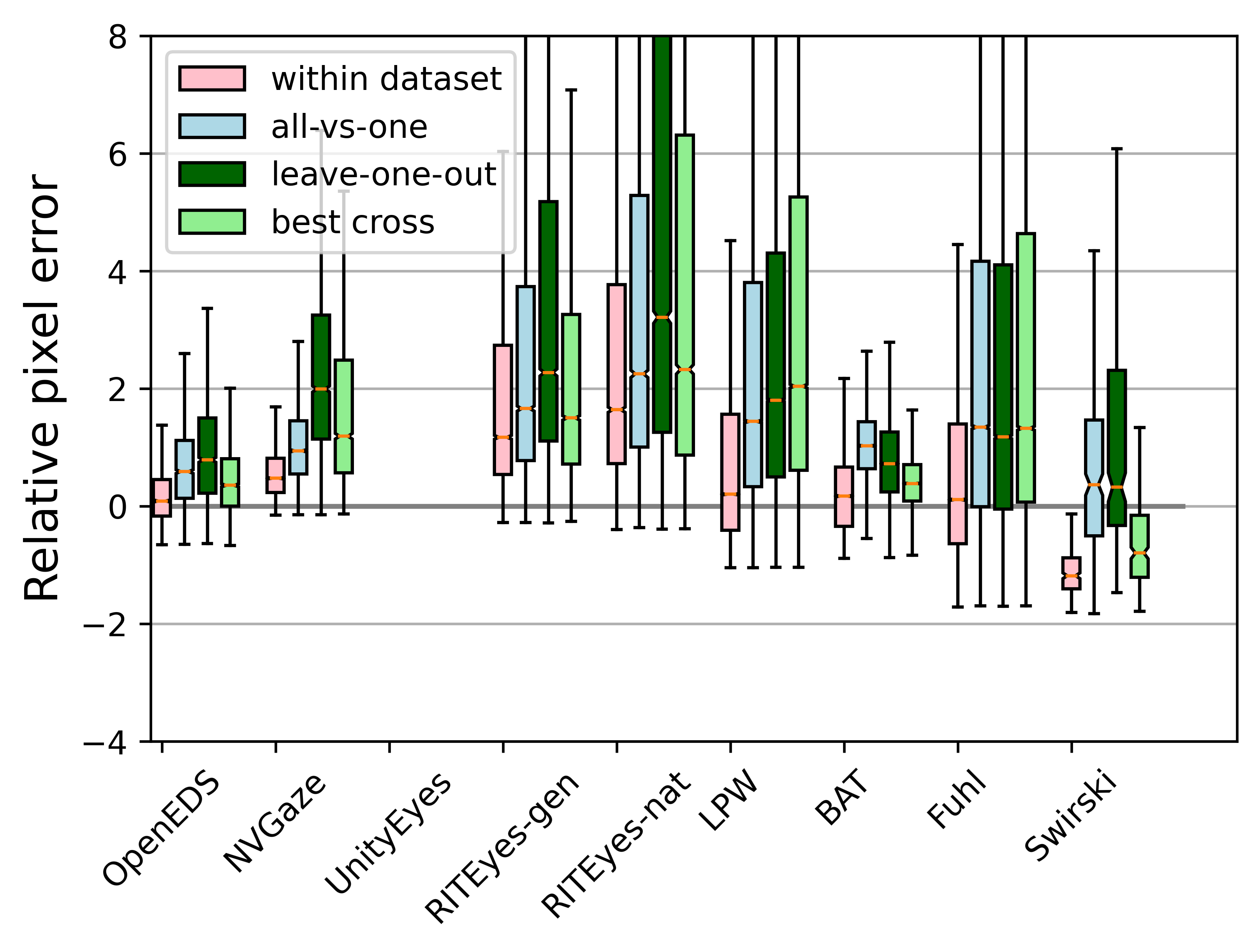}
  \caption{Performance following reduction in network complexity measured as error in pupil center $e_p$ in pixels. Negative values indicate a performance improvement.}
\end{subfigure}
\caption{The effect of reducing model complexity from 2.2M to 529K parameters. This is achieved by setting the growth factor to 1 and increasing the number of grouped filters to 32 per convolution layer. Each box plot highlights a model’s performance centered to the within-dataset limit for each domain. The line and notch present within each box plot represents the median and confidence interval respectively while the ends of each box denotes the 1$^\mathrm{st}$ and 3$^\mathrm{rd}$ quartile. All images are 320$\times$240 resolution. Note that boxplots pertaining to datasets without a certain groundtruth annotation are missing. All measures are centered to the within-dataset threshold as reported in Figure 2 and Supplementary Tables 2, 3 and 4}
\label{fig:gen_results_red_model_comp}
\end{figure}

\end{document}